%% file: main.tex
\documentclass[letterpaper, 10 pt, conference]{ieeeconf}
\IEEEoverridecommandlockouts        
\usepackage{amsmath} %
\usepackage{graphicx}
\usepackage{multicol}
\usepackage[bookmarks=true]{hyperref}

\usepackage{multirow}
\usepackage{booktabs}
\usepackage{todonotes}
\let\labelindent\relax
\usepackage{enumitem}

\usepackage{algorithm}
\usepackage[noend]{algpseudocode}
\usepackage[export]{adjustbox}
\algnewcommand\algorithmicdeclare{\textbf{Assume:}}
\algnewcommand\Declare{\item[\algorithmicdeclare]}

\newcommand{\id}[1]{{\it #1}}
\newcommand{\proc}[1]{\textsc{#1}}
\newcommand{\kw}[1]{\textbf{#1}}

\newcommand{\tamp}{{\sc tamp}}
\newcommand{\pddlstream}{{\sc pddl}stream}

\newcommand{\pred}[1]{\pddl{#1}}
\newcommand{\val}[1]{\pddl{#1}}

\newcommand{\pddl}[1]{{\texttt{#1}}} %
\newcommand{\pddlsmall}[1]{{\small \texttt{#1}}} %

\newcommand{\pddlkwsmall}[1]{\textbf{\pddlsmall{#1}}}

\newcommand{\name}{{\sc M0M}}
\newcommand{\robot}{\mathcal{R}}
\newcommand{\goal}{\mathcal{G}}
\newcommand{\image}{I}
\newcommand{\dbscan}{{\sc dbscan}}
\newcommand{\uois}{{\sc uois}-net-3{\sc d}}
\newcommand{\urlsmall}[1]{{\small \url{#1}}}
\newcommand{\urdf}{{\sc urdf}}

\usepackage{listings}
\title{\LARGE \bf
Long-Horizon Manipulation of Unknown Objects \\ via Task and Motion Planning with Estimated Affordances
}
\author{Aidan Curtis* \and Xiaolin Fang* \and Leslie Pack Kaelbling \and Tom\'as Lozano-P\'erez \and Caelan Reed Garrett%
\thanks{
$^{*}$The first two authors contributed equally and are listed in alphabetical order.
The authors are at CSAIL, MIT, USA:
        {\tt\small \{curtisa, xiaolinf,lpk, tlp, caelan\}@mit.edu}.} 
}
\begin{document}
\maketitle
\thispagestyle{plain}
\pagestyle{plain}

\begin{abstract}
We present a strategy for designing and building very general robot manipulation systems involving the integration of a general-purpose task-and-motion planner with engineered and learned perception modules that estimate properties and affordances of unknown objects.
Such systems are closed-loop policies that map from RGB images, depth images, and robot joint encoder measurements to robot joint position commands.
We show that following this strategy a task-and-motion planner can be used to plan intelligent behaviors even in the absence of a priori %
knowledge regarding the set of manipulable objects, their geometries, and their affordances.
We explore several different ways of implementing such perceptual modules for segmentation, property detection, shape estimation, and grasp generation.  
We show how these modules are integrated within the PDDLStream task and motion planning framework.
Finally, we demonstrate that this strategy can enable a single system to perform a wide variety of real-world multi-step manipulation tasks, generalizing over a broad class of objects, object arrangements, and goals, without any prior knowledge of the environment and without re-training.
\end{abstract}

\section{Introduction}

Our objective is to design and build robot policies that can interact robustly and safely with large collections of objects that are only partially observable, where the objects 
have never been seen before
and where achieving the goal may require many coordinated actions, as in putting away all the groceries or collecting all the ingredients for a meal.
Our goal is a policy that will generalize without specialized re-engineering or re-training to a broad range of novel objects, physical environments, and goals, but also be able to acquire whole new competencies, cumulatively, through incremental engineering and learning.

There is a broad appreciation of the importance of generality in design methods: trajectory optimization and reinforcement learning, for example, are both general tools that can address a large array of problems.  However, the policies that are typically built with them are quite narrow in their domain of application.  We seek, instead, {\em systems generality}, in which the focus is on the generality of a {\em single policy}. In this paper, we describe an approach for building such policies as {\em deliberative systems} and then instantiate it with an implementation that is able to manipulate novel objects in novel arrangements to achieve novel goals, both in simulation and on a real robot.
It makes use of engineered as well as machine-learned modules for object segmentation, shape estimation of 3D object meshes, and grasp prediction, along with a state-of-the-art task-and-motion planner.

\begin{figure}
    \centering
    \includegraphics[trim={2cm 0.5cm 2cm 1cm},clip,width=.49\linewidth]{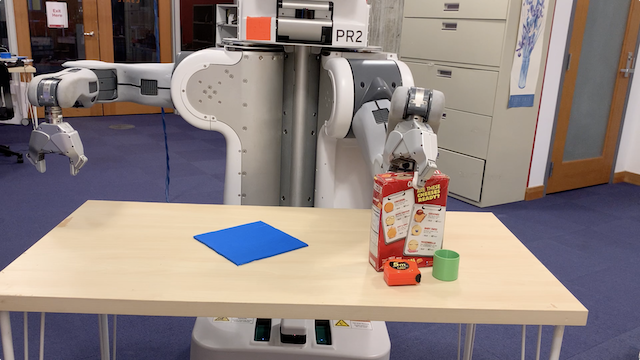}
    \includegraphics[trim={2cm 0.5cm 2cm 1cm},clip,width=.49\linewidth]{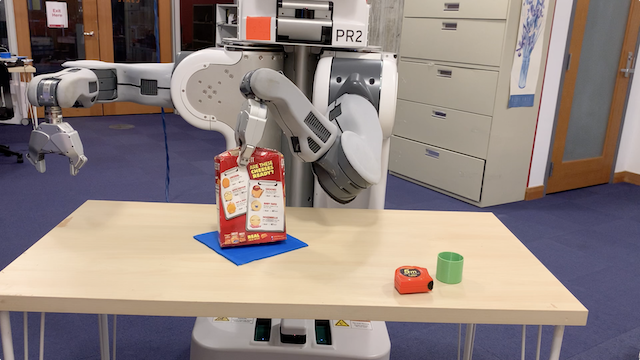} \\
    \vspace{4pt}
    \includegraphics[trim={2cm 0.5cm 2cm 1cm},clip,width=.49\linewidth]{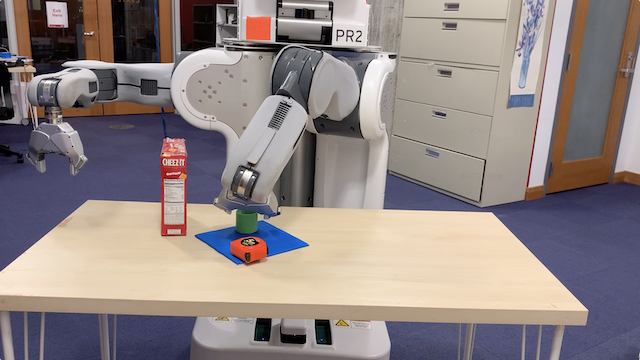}
    \includegraphics[trim={2cm 0.5cm 2cm 1cm},clip,width=.49\linewidth]{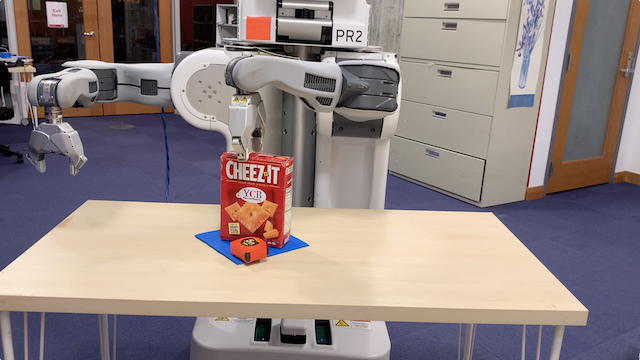}
    \caption{The goal is for all perceivable objects to be on a blue target region. 
    The robot first finds and executes a plan that picks and places the cracker box on the blue target region.
    After re-perceiving the scene and identifying two new objects, the robot finds a joint set of three collision-free placements for all three objects on the target region and plans to safely move them there. 
    }
    \label{fig:task3_intro}
\end{figure}

The operation of our %
system, called \name{} (Manipulation with Zero Models), is illustrated in Figure~\ref{fig:task3_intro}.
The goal is for all objects to be on a blue target region.  {\em Importantly, the system has no prior geometric models of objects and no specification of what objects are present in the world.}  It takes as input RGB-D images, which it segments and processes to find surfaces, colored target regions and object candidates (see Section~\ref{sec:segment}).  The goal for this task is communicated to the system by the following logical formula:
\begin{align*}
\forall {\it obj}.\; \exists {\it region}.\; \pred{On}({\it obj},{\it region}) \wedge \pred{Is}({\it region}, \val{blue}).
\end{align*}
This formula involves a relation \pred{On} that the system knows how to achieve, by picking and placing, and perceptual properties (\pred{Is}) such as color %
that the system can compute from the input images (see Section~\ref{sec:formulation}).
The goal does {\em not} reference any individual objects by name because, in our problem setting, the object instances have no names.
Instead, goals existentially and universally quantify over the perceivable objects, which may vary substantially in number and properties across, and even within, problem instances.

Initially, two objects are purposefully hidden behind the tall cracker box so that the robot cannot perceive them. 
Finding only a single object on the table, the robot first picks and places the cracker box on the blue target region.  It selects a placement for the cracker box on the blue region that is roughly in the middle of the region.
Because the initial cracker box placement was planned without knowledge of the other two objects, upon observing the new objects, the robot intentionally moves the cracker box to a temporary new placement to make room for the tape measure and green cup. 
Finally, the robot plans a new placement for the cracker box that avoids collisions with the other two objects while also satisfying the goal.
A video of this trial is available at: \urlsmall{https://youtu.be/f-GCKQWuPyM}; additional experiments are described in Section~\ref{sec:system-experiments} and in Appendix~\ref{app:demos}.

\name{} can perform purposeful manipulation for a general class of object shapes, object arrangements and goals, while operating directly from perceptual data, even in partially observable settings.  Importantly, the system is designed in a modular fashion so that different modules can be used for perceptual tasks such as segmenting the scene or choosing grasps on detected objects.  Furthermore, new manipulation operations, such as pushing or pouring, can be added and immediately combined with existing operations to achieve new goals.
Many more examples of \name{} in operation are illustrated in the remainder of the paper and at  \urlsmall{https://tinyurl.com/open-world-tamp}.

\section{Approach} \label{sec:approach}

Any robot system that has an extended interaction with its environment, selecting actions based on the world state and the outcomes of its previous actions, can be seen at the most basic level as a control policy that maps a sequence of inputs (generally intensity and depth images and joint angles) into motor torques.  It has been traditional to hand-design and implement such control policies.  A classical strategy of multi-level model-predictive control with a general-purpose planner at the top level results in very robust behavior that generalizes over a wide range of situations and goals
~\cite{Kaelbling2011HierarchicalTA,Srivastava2014CombinedTA,Toussaint2015LogicGeometricPA,Dantam2018AnIC,PDDLStream2020}; however, 
these approaches have traditionally required a substantial amount of prior knowledge of the objects in the world and their dynamics.

A relatively newer strategy for constructing such control policies is to learn them via supervised, imitation, or reinforcement-learning methods in simulation or real-world settings~\cite{Andrychowicz2020LearningDI,Nagabandi2019DeepDM}.  These approaches are attractive because they require less
human engineering %
but they 
make heavy demands for real-world training, which again poses a substantial development burden.  In addition, these learned policies are often narrowly focused on a single ``task''.

In this paper, we present a strategy for obtaining the best of both worlds:  we encode fundamental, very generic, aspects of physical manipulation of objects in three-dimensional space in an algorithmic framework that implements a feedback control policy mapping sensory observations and a goal specification into motor controls.   To instantiate this framework for a new domain one must provide:
\begin{itemize}
    \item A description of the {\em robot's kinematics} and a basic position trajectory controller;
    \item A characterization of the {\em manipulation operations} that robot can use; and 
    \item A set of {\em perceptual modules} 
    that estimate properties of objects that the system will interact with, which can generally be acquired via off-line training and shared over a variety of applications and robots.
\end{itemize}
Systems that instantiate this framework will {\em immediately} generalize without any re-engineering or re-training to a broad range of novel objects, physical environments, and goals.  Due to the modularity of the architecture, they can also serve as a basis for acquiring whole new competencies, cumulatively, by adding new learned or engineered modules.

Our approach leverages the planning capabilities of general-purpose task and motion planning (\tamp{}) systems~\cite{tampSurvey}. 
The key insight behind our approach is that such planners do not necessarily need a perfect and complete model of the world, as is often assumed; they only need answers to some set of ``queries'', which can be answered by direct recourse to perceptual data.  
Existing \tamp{} systems that have been demonstrated in real-world settings, including our prior work,
require known object instance 3D mesh models that can be accurately aligned to the observed data using human-calibrated fiducials or pose estimators, which restricts their applicability to known environments, often with only a few unique object instances~\cite{Kaelbling2011HierarchicalTA,Srivastava2014CombinedTA,Toussaint2015LogicGeometricPA,Dantam2018AnIC,PDDLStream2020}.
Even several extensions to \tamp{}-based systems that actively deal with some uncertainty from perception (such as substantial occlusions)~\cite{Kaelbling2013IntegratedTA,HadfieldMenell2015ModularTA,Garrett2019OnlineRI} require observations in the form of poses of known objects.
This pose registration process is critical for these approaches for identifying human-annotated affordances, such as grasps and placement surfaces, and representing collision volumes during planning.
However, we show that one can also fulfill these operations 
using only the observed point cloud, without the need for prior models. In this paper, we develop a strategy in which all such queries are resolved in sensory data, see Section~\ref{sec:pddlstream}.

The system instance described in this paper 
constructs a ``most likely'' estimate of the current scene by segmenting it into objects that can then be used to estimate 
shapes, grasp affordances, and other salient properties.  It then solves for a multi-step motion plan to achieve the goal given that interpretation, executes the first few steps of the plan, re-observes the scene, determines whether the goal is satisfied, and if not, re-plans.

We demonstrate, in simulation and on a real robot, that our system can handle objects of unknown types and a variety of goals.  
Even if it makes perceptual errors, which are often reflected in taking imperfect actions, it recovers from these problems by continually re-perceiving and re-planning.  We experiment with different implementations of perceptual modules, illustrating the importance of modularity for the overall flexibility and extensibility of the system.

\section{Related work}

We have already discussed relevant work in \tamp{} and in policy learning in Section~\ref{sec:approach}. 
We discuss relevant work in standalone perception in Section~\ref{sec:implementation}.

The most closely related work to ours in manipulation without shape models is by Gualtieri {\it et al.}~\cite{Gualtieri2021RoboticPW}.  Many of the components of our system, {\it e.g.}, grasping and shape estimation, are analogous to theirs.  They, however, assume a task-specific rearrangement planner is provided and do not consider tasks that may require more general manipulation of the environment, {\it e.g.} moving an object out of the way, or the more complex goals enabled by a \tamp{} system.  

A number of other approaches~\cite{Gualtieri2018PickAP,Mitash2020TaskDrivenPA}
demonstrate systems that exploit the ability to gain information by
interacting with objects.  There is also a long line of work aimed at ``interactive segmentation'',
that is, using robot motions to disambiguate
among object hypotheses when manipulating in clutter~\cite{Patten2018ActionSF}.
Object search under partial observability has been studied within a partially observable Markov decision process (POMDP) framework~\cite{Wong2013ManipulationbasedAS,Li2016ActTS}, including work that learns policies that uncover hidden objects in piles~\cite{Kurenkov2020VisuomotorMS}.

\section{Manipulation with zero models}
\label{sec:formulation}

\begin{figure*}
\includegraphics[width=\linewidth]{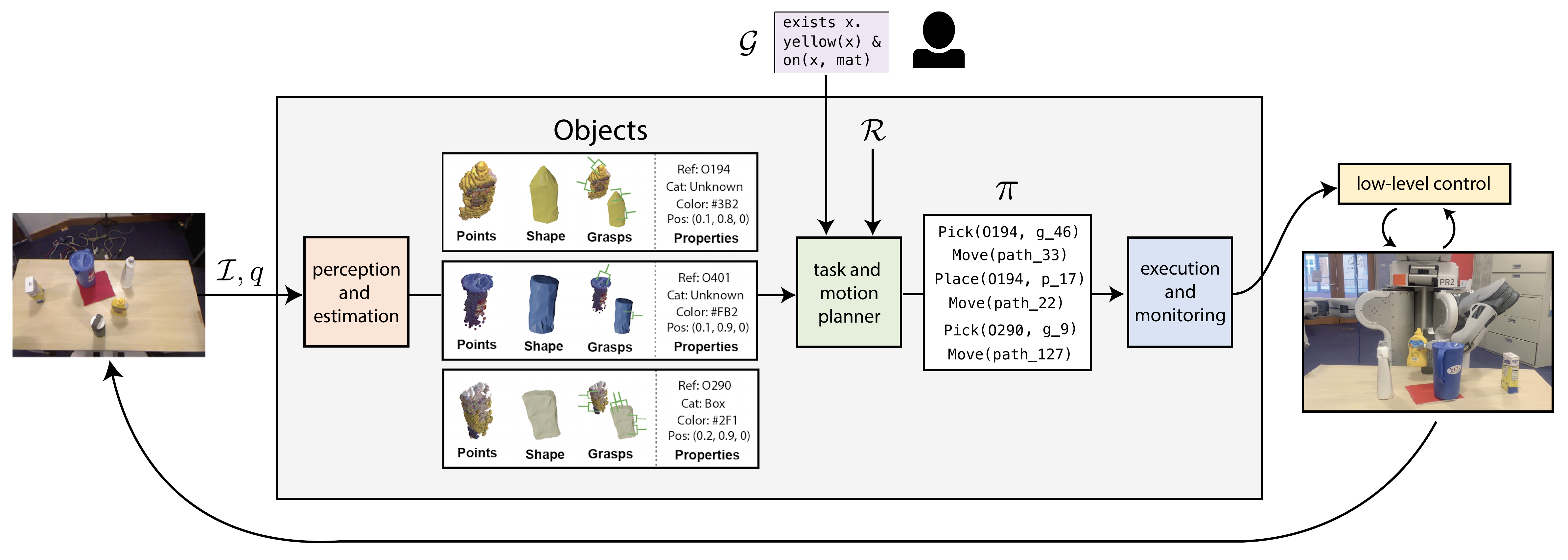}
\caption{Structure of the \name{} policy, which maps RGB-D and robot configuration observations to robot position controls.}
\label{fig:abstractArch}
\end{figure*}

We begin by describing
the scope
of the Manipulation with Zero Models (\name{}) framework for prehensile manipulation, in which the robot moves objects using pick and place operations.   We have previously implemented a variety of other manipulation operations, including pushing, pouring, scooping, and unscrewing bottle caps~\cite{Wang2020LearningCM,Holladay2021Planning}.  
In this paper, we focus on a single, prehensile, manipulation ``mode", which is to pick up objects in a rigid grasp, move them while not contacting any other objects, and then place them stably back onto a surface.  The \name{} system has already built into it the necessary descriptions of these operations for planning; an overview can be found in Section~\ref{sec:pddlstream}.  This single domain description is used for all the objects, arrangements and goals.
The description provided here is the most basic version of the framework;  Section~\ref{sec:extensions} discusses the simplifications and assumptions inherent in this version and outlines strategies for relaxing them.

\paragraph{Robot-specific information} 
To apply \name{} to a manipulation robot, it is necessary to provide a \urdf{} description of the robot's kinematics $\robot{}$ and a position configuration controller for the robot.
The robot may have multiple manipulators that move sequentially.

\paragraph{Perceptual modules}  An instantiation of \name{} requires perceptual modules of several different types.
The first modules take an {\sc RGB-D} image as input:
\begin{description}
\item{\bf rigid objects}: %
Output is a set $O = \{o_1, \ldots, o_{n_o}\}$ of object hypotheses, each of which is characterized by an RGB partial point cloud.  
\item{\bf fixed surfaces}: %
Output is a set $S = \{s_1, \ldots, s_{n_s}\}$ of approximately horizontal surfaces (such as tables, shelves, parts of the floor) that could serve as support surfaces for placing objects.
\end{description}

\noindent{}
Associated with each entity is an arbitrary reference coordinate frame, the simplest being the robot's base frame.
When we speak of $\pred{Pose}(o, p)$, we mean a transform $p$ relative to the reference coordinate frame of $o$.  
This notion of a pose is useful for 
representing relative transformations
but has no semantics outside the system.
The remaining modules take an object point cloud $o$ as input:
\begin{description}
\item{\bf grasps}: 
Output %
is a possibly infinite sequence of transforms between the robot's hand and the reference coordinate frame of $o$ such that, if the robot were to reach that relative pose with the gripper open and then close it, it would likely acquire a stable grasp of $o$.
\item{\bf collision volume}: %
Output is a predicted volume regarding $o$, primarily used for reasoning about collision-avoidance and containment. 
\item{\bf stable orientations}: 
Output is a set of stable orientation of $o$ in its reference frame.
\item{\bf object properties}: 
Output %
is a list of properties of $o$, which will be used in goal specifications.
They can include object class, aspects of shape, color, {\it etc.}
\end{description}
These modules can (and do) use different representations for their computations.  Some may use conservative over-estimates of the input point cloud to find volumes for avoiding collisions, others may use tight approximations of local areas to find candidate grasps, while others may use learned networks operating on the whole input to compute such affordances.

\section{\name{} using \pddlstream{}}  \label{sec:pddlstream}

Our implementation of \name{} uses \pddlstream{}~\cite{PDDLStream2020}, an existing open-source domain-independent planning framework for hybrid domains.  We have previously used \pddlstream{} to solve a rich class of observable manipulation problems;
however, in our previous applications, object shapes were assumed to be known exactly. 
Other \tamp{} frameworks that provide a similar interface between perceptual operations and the planner through for example, suggestors~\cite{Kaelbling2011HierarchicalTA} or a refinement layer~\cite{Srivastava2014CombinedTA}, could also be used as the basis of an implementation of our approach.

\subsection{Overview}

\pddlstream{} takes as input models of the manipulation operations, in the form of Planning Domain Definition Language ({\sc pddl}) operator descriptions (see Figure~\ref{fig:actions}), and a set of {\em samplers} (referred to as {\em streams}), which produce candidate values of continuous quantities, including joint configurations, grasps, object placements, and robot motion trajectories that satisfy the constraints specified in the vocabulary of the problem (see Figure~\ref{fig:streams}).
Critically, aside from a small declaration of the properties that their inputs and outputs satisfy, the implementation of each stream is treated as a {\em blackbox}.
As a result, \pddlstream{} is agnostic toward both the representation of stream inputs and outputs as well as whether operations are implemented using engineering or learning techniques.
This allows state-of-the-art machine learning methods
to be flexibly incorporated, without modification, during planning where they will be automatically combined with other independent operations by the \pddlstream{} planning engine.

Problems described using the \pddlstream{} planning language can be solved by a variety of \pddlstream{} planning algorithms~\cite{PDDLStream2020}.
Because the \pddlstream{} planning engine is responsible for querying the perceptual operations (in the form of streams), it will automatically decide online which operations are relevant to the problem and how many generated values are needed.
Furthermore, several \pddlstream{} algorithms ({\it e.g.} the \textsc{focused} algorithm) will lazily query the perceptual operations in order to avoid unnecessary computation.
As a result, the planner will not perform computationally expensive perceptual operations on images and point clouds to, for example, predict properties and grasps unless the segmented object or property have been identified as relevant to the problem.

Below we describe our \pddlstream{} formulation in detail so as to make the contract between perceptual operations and action descriptions precise.

\begin{figure*}[h!]
\begin{footnotesize}
\begin{lstlisting}
(:action move
 :parameters (?q1 ?t ?q2)
 :precondition (and (Motion ?q1 ?t ?q2) (HandEmpty) (AtConf ?q1)
                    (forall (?oc2 ?p2) (imply (AtPose ?oc2 ?p2) |\underline{(CFreeTrajPose ?t ?oc2 ?p2)}|)))
 :effect (and (AtConf ?q2) (not (AtConf ?q1)))
(:action place
 :parameters (?q ?oc ?g ?p ?oc2 ?p2)
 :precondition (and |\underline{(Grasp ?oc ?g)}| (Kin ?q ?g ?p) |\underline{(Stable ?oc ?p ?oc2 ?p2)}|
                    (AtConf ?q) (AtGrasp ?oc ?g) (AtPose ?oc2 ?p2))
 :effect (and (HandEmpty) (AtPose ?oc ?p) (On ?oc ?oc2) (not (AtGrasp ?oc ?g))))
\end{lstlisting}
\end{footnotesize}
\caption{A \pddlstream{} description of \pddlsmall{move} and \pddlsmall{place} actions. The underlined predicates denote properties estimated by perceptual modules. See Figure~\ref{fig:actions_all} in Appendix~\ref{app:pddlstream} for a description of \pddlsmall{move-holding} and \pddlsmall{pick} actions.
} \label{fig:actions}
\end{figure*}

\subsection{\pddlstream{} formulation} \label{sec:actions}

In {\sc pddl}, an action is specified by a list of free parameters (\pddlkwsmall{:parameters}), a precondition logical formula (\pddlkwsmall{:precondition}) that must hold to correctly apply the action, and an effect logical conjunction (\pddlkwsmall{:effect}) that describes changes to the state when the action is applied.
Figure~\ref{fig:actions} gives the {\sc pddl} description of the \pddlsmall{move} and \pddlsmall{place} actions for \name{}.
The \pddlsmall{move} action models collision-free motion of the robot while it is not holding anything.
The \pddlsmall{place} action models the instantaneous change from when its hand is exerting a force to hold an object to when a force is no longer applied and the object is released.
The \pddlsmall{move-holding} and \pddlsmall{pick} actions are described in Figure~\ref{fig:actions_all} in Appendix~\ref{app:pddlstream}.

A state is a goal state if the goal formula holds in it.
Goal specifications, even those with quantifiers, can be directly and automatically encoded in a {\sc pddl} formulation using {\em axioms}, logical inference rules~\cite{pednault1989adl,thiebaux2005defense,PDDLStream2020}.
Intuitively, an axiom has the same precondition and effect structural form as an action but is automatically derived
at each state.
Due to their similarities with actions, axioms can straightforwardly be incorporated in {\sc pddl}, enabling a planner to efficiently reason about complex goal conditions, such as the ones present in \name{}.

\pddlstream{} builds on {\sc pddl} by introducing stream descriptions, which are similar in syntax to {\sc pddl} operator descriptions.
An stream is declared by a list of input parameters (\pddlkwsmall{:inputs}), a logical formula that all legal input parameter values {\em must} satisfy (\pddlkwsmall{:domain}), a list of output parameters (\pddlkwsmall{:outputs}), and a logical conjunction that all legal input parameter values and generated output parameter values are {\em guaranteed} to satisfy (\pddlkwsmall{:certified}).
Each stream is accompanied by a procedure that maps input parameter values to a possibly infinite sequence of output parameter values.
Figure~\ref{fig:streams} displays six streams used in \name{}, which we will describe in detail in Section~\ref{sec:streams}.

\subsection{Predicate semantics} \label{sec:predicates}

First, we describe the semantics of the predicates used in Figure~\ref{fig:actions} and Figure~\ref{fig:streams}.
The following predicates to encode parameter values {\em type}:
\pddlsmall{(Conf ?q)} indicates \pddlsmall{?q} is a continuous robot joint configuration;
\pddlsmall{(Traj ?t)} indicates \pddlsmall{?t} is a continuous robot joint trajectory;
\pddlsmall{(ObjectCloud ?oc)} indicates \pddlsmall{?oc} is an object, which crucially is represented by a segmented point cloud observation; %
\pddlsmall{(Pose ?oc ?p)} indicates \pddlsmall{?p} is a pose transform for an object point cloud \pddlsmall{?oc} relative to its observed frame; %
\pddlsmall{(Grasp ?oc ?g)} indicates \pddlsmall{?g} is a grasp transform for an object point cloud \pddlsmall{?oc} relative to its observed frame.
The choice to use the observed frame as the reference frame for an object is arbitrary and has no bearing on the system as poses are only used internally during planning.
As a result of this decision, the initial pose $p_0$ of each object cloud is the identity pose.
\pddlsmall{(Property ?pr)} denotes that \pddlsmall{?pr} is a perceivable property, such as a particular color or category.

The following {\em fluent} predicates model the current state of the system: \pddlsmall{(AtConf ?q)} represents the current robot configuration \pddlsmall{?q}; \pddlsmall{(HandEmpty)} is true if the robot's hand is empty; \pddlsmall{(AtGrasp ?oc ?g)} indicates that object cloud \pddlsmall{?oc} is held by the robot at grasp \pddlsmall{?g}; \pddlsmall{(AtPose ?oc ?p)} indicates that object cloud \pddlsmall{?oc} is resting at placement \pddlsmall{?p}; \pddlsmall{(On ?oc ?oc2)} signifies that object cloud \pddlsmall{?oc} is resting on object cloud \pddlsmall{?oc2}.
Normally, in a fully observable \tamp{} setting, \pddlsmall{?oc} would be the name of an object instance; however, those do not exist in our setting, so \pddlsmall{?oc} is simply a unique point cloud.
The initial planning state of the system after $n$ object clouds $\pddlsmall{oc\#1}, ..., \pddlsmall{oc\#n}$ were segmented from the last observation is:
\begin{align*}
    s_0 = \{&\pddlsmall{AtConf}(q_0), \pddlsmall{HandEmpty}(), \\ &\pddlsmall{AtPose}(\pddlsmall{oc\#1}, p_0^1), ..., \pddlsmall{AtPose}(\pddlsmall{oc\#n}, p_0^n)\}
\end{align*}
where $q_0$ is the current robot configuration, $p_0^1, ..., p_0^n$ are identity poses, and the robot's hand is empty.

\subsection{Engineered and learned streams} \label{sec:streams}

\begin{figure*}[h!]
\begin{footnotesize}
\begin{multicols}{2}
\begin{lstlisting}
(:stream predict-grasps
 :inputs (?oc)
 :domain (ObjectCloud ?oc)
 :outputs (?g)
 :certified |\underline{(Grasp ?oc ?g)}|)
(:stream inverse-kinematics
 :inputs (?oc ?p ?g)
 :domain (and (Grasp ?oc ?g) (Pose ?oc ?p))
 :outputs (?q)
 :certified (and (Kin ?q ?g ?p) (Conf ?q)))
(:stream plan-motion
 :inputs (?q1 ?q2)
 :domain (and (Conf ?q1) (Conf ?q2))
 :outputs (?t)
 :certified (and (Motion ?q1 ?t ?q2) (Traj ?t)))
(:stream predict-placements
 :inputs (?oc1 ?oc2 ?p2)
 :domain (and (ObjectCloud ?oc1) (Pose ?oc2 ?p2))
 :outputs (?p1)
 :certified (and |\underline{(Stable ?oc1 ?p1 ?oc2 ?p2)}| 
                 (Pose ?oc1 ?p1)))
(:stream predict-cfree
 :inputs (?t ?oc2 ?p2)
 :domain (and (Traj ?t) (Pose ?pc2 ?p2))
 :certified |\underline{(CFreeTrajPose ?t ?oc2 ?p2)}|)
(:stream detect-property
 :inputs (?oc ?pr)
 :domain (and (ObjectCloud ?oc) (Property ?pr))
 :certified |\underline{(Is ?oc ?pr)}|)
\end{lstlisting}
\end{multicols}
\end{footnotesize}
\caption {A \pddlstream{} description of the streams, 
which represent engineered and learned operations.
The underlined predicates denote properties estimated by perceptual modules.  
See Figure~\ref{fig:streams_all} in Appendix~\ref{app:pddlstream} for a description of the \pddlsmall{predict-traj-grasp} stream.
} \label{fig:streams}
\end{figure*}

Next, we describe the streams as well as the {\em constraint} predicates that they certify.
We highlight the distinction between streams that can be directly {\em engineered} and those that must be at least partially {\em learned}.
The engineered streams we consider are robot-centric operations that can performed using the robot's fully-observed \urdf{}, which encodes the robot's kinematics and geometry.
The \pddlsmall{inverse-kinematics} stream solves for configurations \pddlsmall{?q} that satisfy the kinematic constraint \pddlsmall{(Kin ?q ?g ?p)} with grasp \pddlsmall{?g} and pose \pddlsmall{?p}, for example, using IKFast~\cite{diankov2010automated}.
The \pddlsmall{plan-motion} stream plans a continuous trajectory \pddlsmall{?t} between configurations \pddlsmall{?q1} and \pddlsmall{?q2} that respects joint limits and self collisions, certifying \pddlsmall{(Motion ?q1 ?t ?q2)}.
It can be directly implemented by any off-the-shelf motion planner, such as RRT-Connect~\cite{KuffnerLaValle}.

The learned streams can use a combination of machine learning and classical estimation techniques.
In our system, we consider several implementations of each stream that each are a wrapper around a state-of-the-art estimation technique for their subproblem.
The \pddlsmall{predict-grasps} stream generates grasps \pddlsmall{?g} for object cloud \pddlsmall{?oc} that are predicted to remain stably in the robot's hand, certifying \pddlsmall{(Grasp ?oc ?g)}.
In Section~\ref{sec:grasp}, we describe several machine learning implementations of \pddlsmall{predict-grasps}, some of which make predictions directly from \pddlsmall{?oc} without any intermediate representation.

The \pddlsmall{predict-placements} stream generates poses \pddlsmall{?p1} for object cloud \pddlsmall{?oc1} that are predicted to rest stably on object cloud \pddlsmall{?oc2} when at pose \pddlsmall{?p2}, certifying \pddlsmall{(Stable ?oc1 ?p1 ?oc2 ?p2)}).
Our implementation of \pddlsmall{predict-placements} decomposes the operation into two estimation subprocedures.
First, we perform point cloud completion (Section~\ref{sec:shape}) on object cloud \pddlsmall{?oc2} and then estimate approximately horizontal planar surfaces in \pddlsmall{?oc2} when at pose \pddlsmall{?p2} using Random Sample Consensus (RANSAC)~\cite{fischler1981random} plane estimation.
Next, we perform shape estimation (Section~\ref{sec:shape}) on object cloud \pddlsmall{?oc1} and then estimate stable orientations relative to a planar surface using the resulting mesh~\cite{goldberg1999part}. %
By combining these two subprocedures, we obtain placements \pddlsmall{?p1} for object cloud \pddlsmall{?oc1}.

The \pddlsmall{predict-cfree} stream predicts whether all robot configurations along trajectory \pddlsmall{?t} do not collide ({\it i.e.} are collision-free) with object cloud \pddlsmall{?oc2} at pose \pddlsmall{?p2}, certifying \pddlsmall{(CFreeTrajPose ?t ?oc2 ?p2)}.
By finely sampling configurations along trajectory \pddlsmall{?t}, this test can be reduced to sequence of robot configuration and object cloud collision predictions.
Although these predictions could be made directly, we instead use shape estimation (Section~\ref{sec:shape}) to estimate the collision volume of both the observable and unobservable object volume as a set of convex bodies.
This enables us to use fast convex body collision checkers to answer these queries~\cite{gjk1988}. %
A similar \pddlsmall{predict-traj-grasp} stream that predicts collisions with a grasped object is described in %
Appendix~\ref{app:pddlstream}.
Finally, the \pddlsmall{detect-property} stream tests whether object cloud \pddlsmall{?oc2} has property \pddlsmall{?pr2} and, if so, certifies \pddlsmall{(Is ?oc ?pr)}.
Section~\ref{sec:property} describes two property estimators, which detect the category and color of an object from the RGB image observation.

\section{Manipulation policy}

The pseudocode for the manipulation policy, which at its core leverages planning using the model described in Section~\ref{sec:pddlstream}, is displayed in Algorithm~\ref{alg:policy}.
The \name{} solution strategy 
A flowchart of the policy is illustrated in Figure~\ref{fig:abstractArch}.
The policy assumes the set of manipulation actions ${\cal A}$ (Section~\ref{sec:actions}) and the engineered streams ${\cal S}_E$ (Section~\ref{sec:streams}).
It requires a implementation of the learned streams ${\cal S}_L$.
Several options per stream are discussed in Section~\ref{sec:implementation}.
The policy is conditioned on a particular robot ${\cal R}$ and a specified goal ${\cal G}$.
To apply the policy to a new robot ${\cal R}$, it is necessary to provide a \urdf{} description of the robot's kinematics $\robot{}$ and a position configuration controller for the robot.

On each decision-making iteration, the robot receives the current RGB-D image $I$ from its camera and its current joint configuration $q$ from its joint encoders.
From each input RGB-D image, it segments out table point clouds $T$ and object point clouds $O$.
The segmented object and table point clouds as well as the robot configuration instantiate the current \pddlstream{} state $s$ of the world and robot.
This current state along with the goal ${\cal G}$, actions ${\cal A}$, and streams ${\cal S}_E \cup {\cal S}_L$ form a \pddlstream{} planning problem, which is solved by \proc{solve-pddlstream}, a procedure that denotes a generic \pddlstream{} planning algorithm.
In some cases, such as when a necessary attribute is not detected, \proc{solve-pddlstream} will return \kw{None}, indicating that the goal ${\cal G}$ is unreachable from the current state $s$.
Otherwise, \proc{solve-pddlstream} will return a plan $\pi$, which consists of a finite sequence of instances of the actions in ${\cal A}$.
If the plan is empty ({\it i.e.} $\pi = [\;]$), the current state $s$ was proved to satisfy the goal and the policy terminates successfully.
Otherwise, the robot executes the first action $a_1 = \pi[0]$ using its position controllers and repeats this process by reobserving the scene.
Note that this control structure enforces that the robot observes the scene to infer whether it has achieved the goal; otherwise, the robot could erroneously declare success after executing a plan open loop.

\begin{algorithm}[!h]
  \caption{The \name{} policy}
  \label{alg:policy}
  \begin{algorithmic}[1] %
    \begin{small}
    \Declare ${\cal A} = \{\pddl{move}, \pddl{move-holding}, \pddl{pick},  \pddl{place}\}$
    \Declare ${\cal S}_E = \{\pddl{inverse-kinematics}, \pddl{plan-motion}\}$
    \Require ${\cal S}_L = \{\pddl{predict-grasps}, \pddl{predict-placements},$
    \Statex $\pddl{predict-cfree}, ...,  \pddl{detect-attribute}\}$
    \Procedure{execute-policy}{$\robot{}, \goal{}$} \text{Robot \urdf{} $\robot{}$, goal $\goal{}$}
	  \While{\kw{True}}
	    \State $\image, q \gets \proc{observe}()$ \Comment{RGB-D image $\image$, robot conf $q$} %
	    \State $T, O \gets \proc{segment}(\image{})$ \Comment{Tables $T$, objects $O$}
	    \State $s \gets \proc{obj-state}(\image{}, T, O) \cup \proc{robot-state}(\robot{}, q)$ %
	    \State $\pi \gets \proc{solve-pddlstream}(s, \goal{}, {\cal A}, {\cal S}_E \cup {\cal S}_L)$ %
	    \If{$\pi = \kw{None}$} 
	         \Return \kw{False} \Comment{Failure: unreachable}
	    \EndIf
	    \If{$\pi = [\;]$}
             \Return \kw{True} \Comment{Success: $s \in {\cal G}$} 
        \EndIf
	    \State \proc{execute-action}($\robot{},\pi[0]$)
       \EndWhile
       \EndProcedure
    \end{small}
\end{algorithmic}
\end{algorithm}

\section{Implementation}
\label{sec:implementation}
We have implemented an instance of \name{}, %
using \pddlstream{} and experimented with different strategies for implementing the perceptual modules.  All make use of RGB-D images gathered from the PR2's Kinect 1 sensor.
In this section, we briefly describe implementation of individual modules and include experimental results comparing alternative implementations of several of the modules.

We used standard position trajectory controllers for simulation in PyBullet~\cite{coumans2019} and on a physical PR2 robot, and simply opened and closed the parallel-jaw grippers to implement grasping and releasing objects.  We used the actual opening of the gripper after commanding the gripper to close to detect grasp failure.

\subsection{Segmentation of objects and surfaces} \label{sec:segment}

\begin{figure*}[h!]
    \centering
    \includegraphics[width=.32\linewidth]{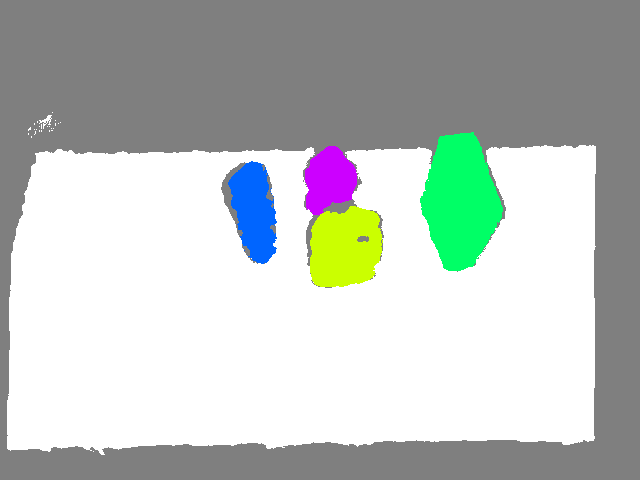}
    \includegraphics[width=.32\linewidth]{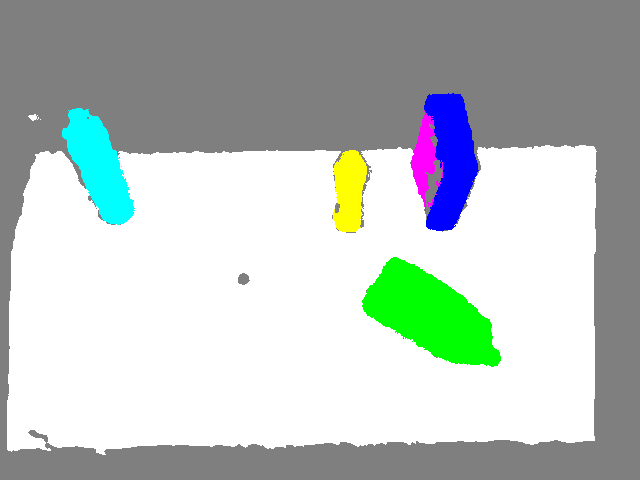}
    \includegraphics[width=.32\linewidth]{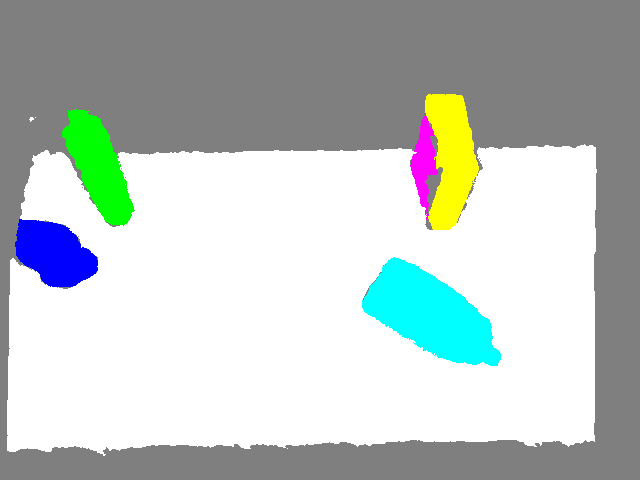}
    \caption{Three segmentation masks predicted by \uois{} during the system's execution in Figure~\ref{fig:task5}. White pixels correspond to the table, chromatically-colored pixels correspond to object instances, and grey pixels are unassigned. Our system does not track objects over time, so each object instance is independently and arbitrarily assigned a color. {\it Left}: the initial segmentation mask. {\it Middle}: an intermediate segmentation mask after picking and placing two objects. {\it Right}: the final segmentation mask in a goal state.}
    \label{fig:observations}
\end{figure*}

\textit{Category-agnostic segmentation} is used to identify rigid collections of points that collectively move as an object when manipulated. %
We compare three different segmentation approaches: \uois{}, geometric clustering, and a combined method.
\uois{}~\cite{xie2020uois3d} is a %
neural-network model that takes RGB-D images as input and returns a segmentation of the scene.  It assumes that objects are generally resting on a table; it attempts to segment out image regions corresponding to the table, as well as a set of objects.

For geometric clustering, we first remove the points assigned to the table by \uois{}, then use {\it density-based spatial clustering of applications with noise} (\dbscan)~\cite{ester1996dbscan}, which finds connected components in a graph constructed by connecting points in the point cloud that are nearest-neighbors in 3D Euclidean distance.
In a combined approach, we apply \dbscan{} to the segmented point cloud produced by \uois{} in order to reduce under-segmentation.  We additionally use post-processing to filter degenerate clusters.

Figure~\ref{fig:observations} displays the segmentation mask predicted by \uois{} while our system was executing the task 5 trial displayed in Figure~\ref{fig:task5}.
As can be seen, \uois{} generally correctly segments the four instances; however, it does oversegment the cracker box into two contacting instances in the last two images.

We compared all three segmentation methods on the ARID-20 subset of the {\it object clutter indoor dataset} (OCID)~\cite{Suchi2019easylabel} and {\it GraspNet-1Billion}~\cite{Fang2020graspnet} datasets.  Detailed results are reported in the appendix.  We found that the different segmentation algorithms have advantages in different settings. In domains where objects have simple geometries and are scattered on the table, an Euclidean-based approach produces reliable predictions.
But in a more cluttered domain, the learned approach often outperforms the Euclidean-based approach.
When it comes to challenging situation where objects have more complicated geometry, the performance of the learned approach drops but it still outperforms the pure Euclidean-based approach. In all experiments, the combined approach performs better than the pure learned approach, indicating the effectiveness of applying \dbscan{} and filtering to neural-network-predicted results.
In our system experiments, we use the combined method.

\subsection{Shape estimation}
\label{sec:shape}

A subroutine of our implementation of both the \pddlsmall{predict-placements} and \pddlsmall{predict-cfree} stream operations (Section~\ref{sec:streams}) is {\em shape estimation}, which takes in a partial point cloud as input and predicts a completed volumetric mesh.
We again explore a combination of neural-network-based and geometric methods.

The {\it morphing and sampling network} (MSN)~\cite{liu2020morphing} is a 
neural-network model that takes as input a partial point cloud and predicts a completed point cloud. 
Our geometric method works by augmenting the partial point cloud by computing the projection of the visible points onto the table plane.  This simple heuristic is motivated by the intuition that the base of an object must be large enough to stably support the visible portion of an object and is particularly useful given a viewpoint that tends to observe objects from above.
As a post-processing step for both methods, we filter the result by back-projecting predicted points onto the depth image and pruning any visible points that are closer to the camera than the observed depth value.

\subsubsection{Mesh interpolation}

While it is possible directly use the estimated point cloud in downstream operations, for example by treating the points as spheres or downsampling them as into voxel grid, it is more accurate and efficient to interpolate among the points to produce a volumetric mesh.
The simplest way to do this is to take the convex hull of the points; however, this can substantially overestimate the volume when the object is non-convex and fail to find feasible plans when attempting to grasp non-convex objects such as bowls.
Instead, we produce the final volume by computing a ``concave hull'' in the form of an alpha shape~\cite{edelsbrunner1983alphashape}, a non-convex generalization of a convex hull, from the union of the visible, network-predicted, and projected points.   To enable efficient collision checking in the \pddlsmall{predict-cfree} stream, we build an additional representation that approximates the mesh as the union of several convex meshes, implemented by {\it volumetric hierarchical approximate convex decomposition} (V-HACD)~\cite{muller2013real}.

\begin{figure*}[h!]
    \centering
    \includegraphics[width=.24\linewidth]{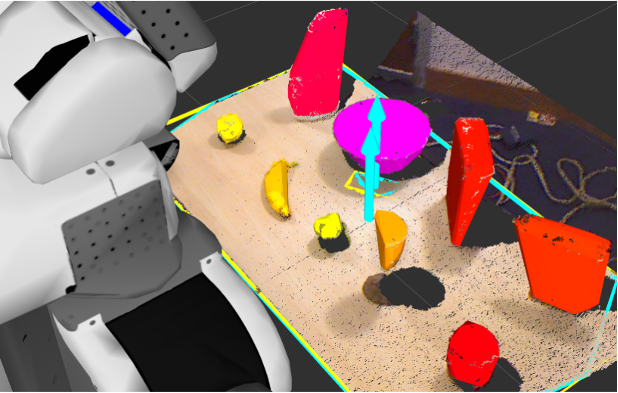}
    \includegraphics[width=.24\linewidth]{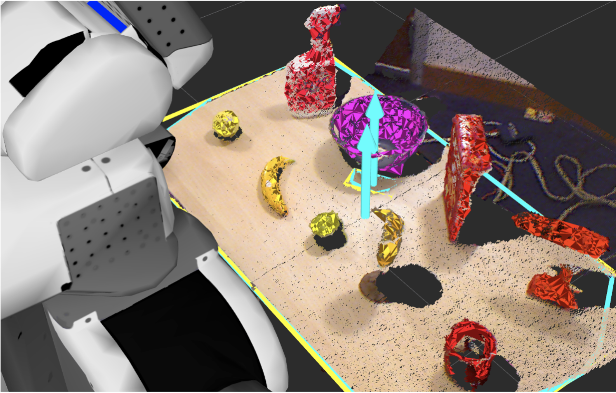}
    \includegraphics[width=.24\linewidth]{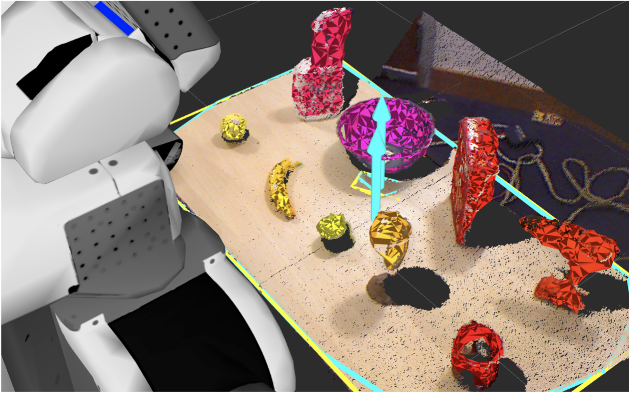}
    \includegraphics[width=.24\linewidth]{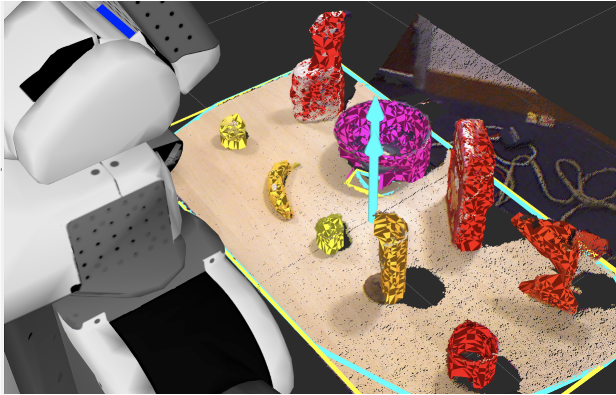}
    \caption{RViz visualizations of estimated shapes overlaid on top of raw point cloud data. a) the convex hull of the visible points only (\textit{V}), b) the concave hull of the visible points only (\textit{V}), c) the concave hull of the visibility-filtered visible and shape-completed points (\textit{VLF}), and d) the concave hull of the visibility-filtered visible, shape-completed, and projected points points (\textit{VLPF}).}
    \label{fig:rviz}
\end{figure*}

Figure~\ref{fig:rviz} visualizes the estimated meshes produced by four of the shape estimation strategies in an uncluttered scene with a diverse set of objects.
The first two images compare creating a mesh by taking the convex hull (Figure~\ref{fig:rviz} {\it left}) versus a concave hull (Figure~\ref{fig:rviz} {\it middle-left}) of the set of visible points (\textit{V}).
The convex hull can significantly overestimate non-convex objects in certain areas, as evidenced by the spray bottle in the top left of the image and the real-world power drill in the right side of the image.
The last three images compare three strategies for populating the set of points to be used the input to a concave hull.
Adding the shape-completed points from MSN (\textit{VLF}) fills in some but not all of the occluded volume of each object, as shown by the cracker box in the middle of the image (Figure~\ref{fig:rviz} {\it middle-right}).
Also including the projection of the points to the table (\textit{VLPF}) better fills in the occluded volume at the cost of overestimating the volume when the ground truth base projection is smaller than the visible base projection (Figure~\ref{fig:rviz} {\it right}).
We evaluated the performance of these methods in four different domains, each on 2000 images taken from a randomly-sampled camera pose; details of the experiments and results can be found in Appendix~\ref{app:perception}.  The fully combined method (\textit{VLPF}) in general performed the best across the domains and is the one we use in the system experiments.

\subsection{Grasp affordances}
 \label{sec:grasp}
Grasp affordances are 
transformations between the robot's hand and an object's reference frame such that, if the robot's hand was at that pose and closed its fingers, it would acquire the object in a stable grasp.
They are purely local and do not take reachability, obstacles, or other constraints into account.
The modularity of our planning framework enables us to consider three interchangeable grasping methods for implementing the \pddlsmall{predict-grasps} stream, each take a partial point cloud as input.
Grasp Pose Detection (GPD)~\cite{Gualtieri2016HighPG} first generates grasp candidates by
aligning one of the robot's fingers to be parallel to an estimated surface in the partial point cloud and then
scores these candidates using a convolutional neural network, which is trained on successful grasps for real objects. 
GraspNet~\cite{mousavian2020graspnet} 
uses a variational autoencoder (VAE) to learn a latent space of grasps that, when conditioned on a partial point cloud, yields grasps. 

We also developed a method, {\em estimated mesh antipodal} (EMA), that performs shape estimation using the methods described in Section~\ref{sec:shape} and then identifies antipodal contact points on the estimated mesh.
Specifically, to generate a new grasp, EMA %
samples two points on the surface of the estimated mesh that are candidate contact points for the center of the robot's fingers.
The pair of points is rejected if the distance between them exceeds the gripper's maximum width or if the surface normal at either of the corresponding faces is not approximately parallel to the line between the two points.
Then, EMA samples a rotation for the gripper about this line and yields the resulting grasp if the gripper, when open and at this transformation, does not collide with the estimated mesh.
A key distinction between EMA's and GPD's candidate grasp generation process is that, by using shape estimation, EMA is able to directly reason about the occluded regions of an object instead of just the visible partial point cloud.
Additionally, it can take into account unsafe contacts between the robot's gripper and the object.

\begin{figure}
    \centering
    \includegraphics[width=\linewidth]{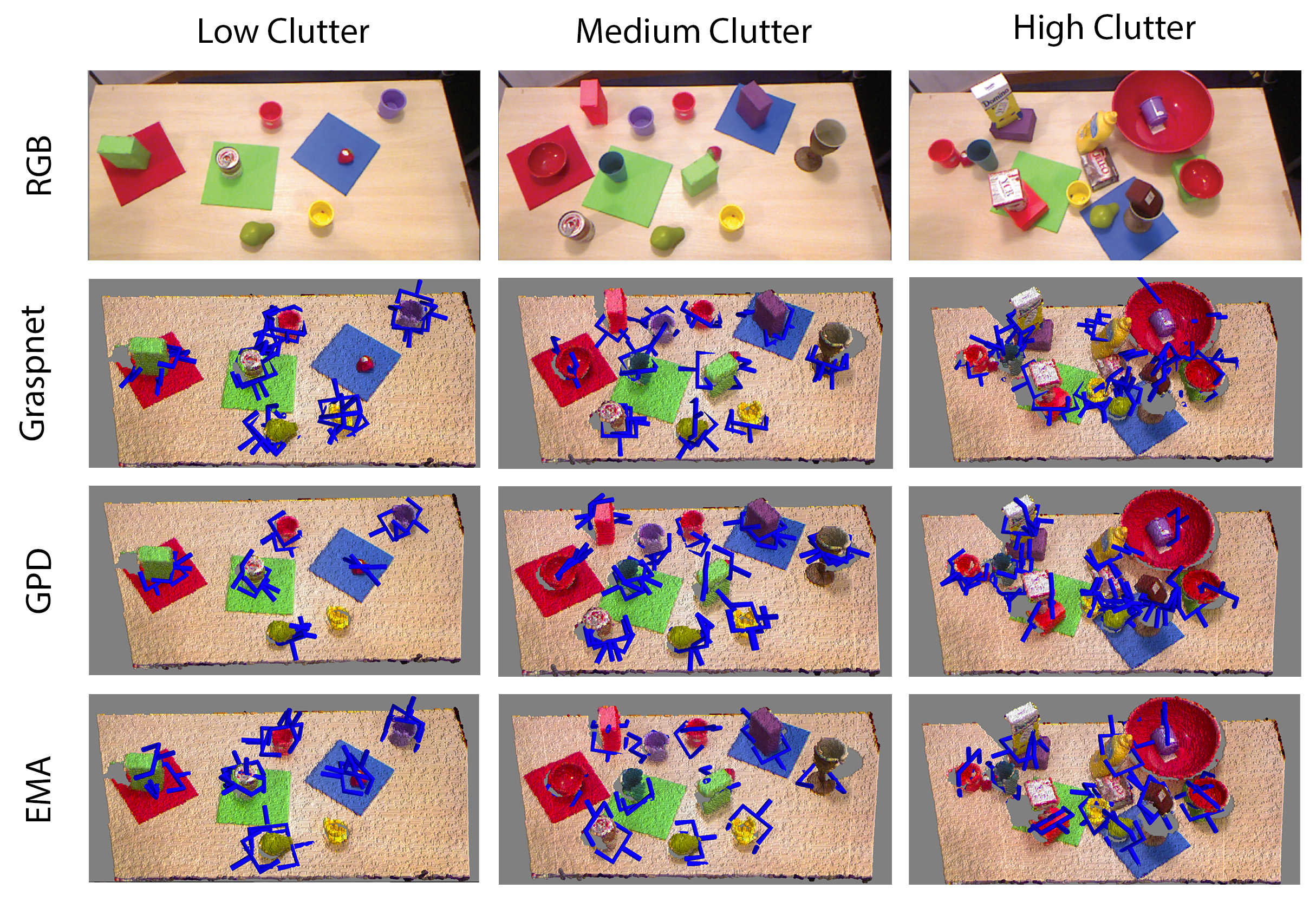}
    \caption{Grasps produced by each of the three grasp generation modules overlaid in green on the observed point cloud for three scenes with varying amounts of clutter.}
    \label{fig:grasp_qualitative}
\end{figure}

Figure~\ref{fig:grasp_qualitative} illustrates some of the grasps produced by these three approaches in three scenes with varying amounts of clutter, where clutter introduces additional opportunities for occlusion.
We performed a real-world experiment to compare the success rates of GPD, GraspNet, and EMA.  The details of the experiment and the results can be found in Appendix~\ref{app:perception}.  GPD and EMA outperformed GraspNet in our experiments, both in speed and accuracy, with EMA having an edge over GPD.  We used EMA in our system experiments.

\subsection{Object properties} \label{sec:property}
In our implementation of the \pddlsmall{detect-property} stream, we considered detectors for two object properties: category and color.  
We use Faster R-CNN~\cite{Ren2016Faster} trained on the \textit{bowl-cup} subset of IIT-AFF~\cite{Nguyen2017IITAffordance} to detect bowls and cups so that the robot can identify which objects can contain other objects.
Additionally, we use Mask R-CNN~\cite{he2017mask}, trained on both real images in Yale-CMU-Berkeley (YCB) Video Dataset~\cite{Xiang2018Posecnn} and a synthetic dataset we generated using PyBullet~\cite{coumans2019}, to classify any YCB objects~\cite{calli2015ycb} that are mentioned in the goal formula.
We also have simple modules that aggregate color statistics directly from segmented RGB images.

\section{Whole-system experiments} \label{sec:system-experiments}

\begin{figure*}[h!]
    \centering
    \includegraphics[trim={1.5cm 0.25cm 2.5cm 1.25cm},clip,width=.32\linewidth]{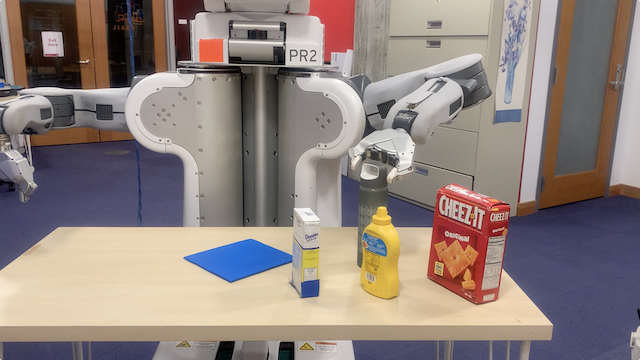}
    \includegraphics[trim={1.5cm 0.25cm 2.5cm 1.25cm},clip,width=.32\linewidth]{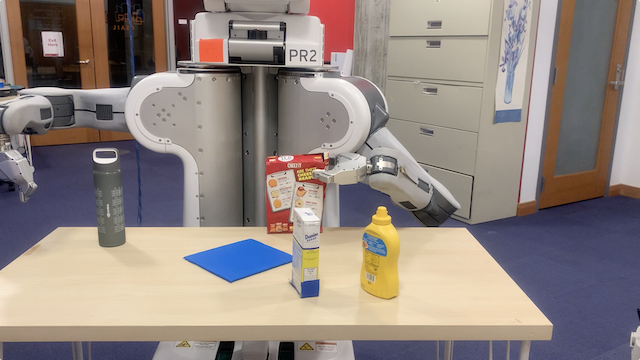}
    \includegraphics[trim={1.5cm 0.25cm 2.5cm 1.25cm},clip,width=.32\linewidth]{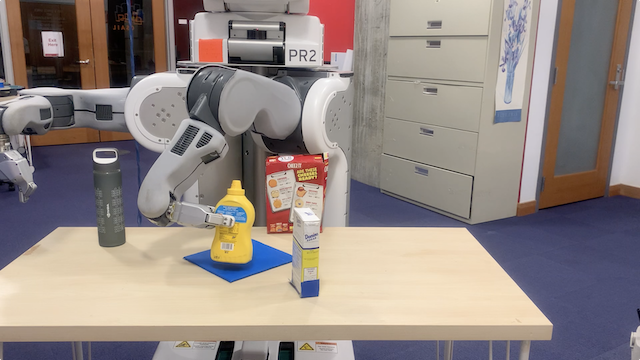}
    \caption{\name{} acting to satisfy the goal for a mustard bottle to be on a blue target region. {\it Left}: for its first action, the robot picks up a water bottle that prevents the robot from safely reaching the mustard bottle. {\it Middle}: the robot also picks and relocates the obstructing cracker box. {\it Right}: finally, the robot places the mustard bottle on the blue target region.}
    \label{fig:obstructions}
\end{figure*}

Finally, we evaluated the whole \name{} system by testing its ability to solve challenging real-world manipulation tasks.
As an example, Figure~\ref{fig:obstructions} illustrates a task where the goal is for a mustard bottle to be on a blue target region:
\begin{align*}
\exists {\it obj}.\; \exists {\it region}.\;  & \pred{On}({\it obj},{\it region}) \wedge \pred{Is}({\it region},\val{blue}) \\
& \wedge \pred{Is}({\it obj},\val{mustard}).
\end{align*}
In its effort to solve this task, the robot moved two obstructing objects out of the way to safely pick the mustard bottle and then place it on the goal region.
Additionally, although not pictured, the robot's first attempt to pick the mustard bottle fails, causing the system to abort execution, re-observe, re-plan, and execute a new grasp that this time was successful.
The full video of the trial can be seen at \urlsmall{https://youtu.be/tNHjpXP8RFo}.

\subsection{Repeated trials}
We performed experiments consisting of five repeated real-world trials for five tasks, obtaining the results shown in Figure~\ref{tab:system}.
Here, tasks are loosely defined as a set of problems with the same goal formula and qualitatively similar initial states.  
Recall that, outside of the PR2's description, the goal formula is the {\it only input} to each trial.
We summarize the results here and describe the tasks in subsequent subsections.  In addition to these tasks, we applied the system to over 25 individual problem instances.
Appendix~\ref{app:demos} highlights several of these tasks along with how the system behaved to solve them.
See \urlsmall{https://tinyurl.com/open-world-tamp} for full videos of our system solving these problems.

The column {\it Iterations} refers to the average number of combined estimation and planning iterations that were performed per trial.
Unless the initial state satisfies the goal conditions, the system always takes two or more iterations because it must at least achieve the goal and then validate that the goal is in fact satisfied.
Sometimes the system will perform more than two iterations in the event that the perception module identifies a new object due to undersegmentation, an action is aborted due to a failed grasp, or an action has unanticipated effects. 

The columns {\it Estimation}, {\it Planning}, and {\it Execution} report the average time spent perceiving, planning, and executing per iteration.
Each module was implemented in Python to flexibly support multiple implementations of each module.
Many of the perceptual operations that manipulate raw point clouds could be sped up by using C++ instead of Python and deploying the system using state-of-the-art graphics hardware.
During planning, a majority of the time is spent checking for collisions, particularly when the robot is planning free-space motions.
The overall runtime could be reduced by simultaneously planning motions for later actions while executing earlier actions~\cite{Garrett2019OnlineRI}.
The column {\it Successes} reports the number of times out of five trials that the system terminated having identified that it achieved the goal.
Our system was able to achieve the goal on every trial except for a single trial that was a part of {\it Task 3}. These results show that this single system can perform a diverse set of long-horizon manipulation tasks robustly and reliably.

\begin{figure}[h]
\centering \small
\begin{tabular}{ cccccc } \toprule %
{Task} &
{Iterations} & %
{Estimate} & %
{Plan} & %
{Execute} & %
{Success} \\ %
\midrule
1 & 2.0 & 29.6s & 18.5s & 16.1s & 5/5 \\ 
2 & 3.0 & 34.0s & 37.4s & 23.6s & 5/5 \\ 
3 & 3.0 & 36.8s & 28.3s & 40.5s & 4/5 \\ 
4 & 2.0 & 39.6s & 41.6s & 44.6s & 5/5 \\ 
5 & 2.4 & 47.7s & 18.9s & 28.3s & 5/5 \\
\bottomrule
\end{tabular}
\caption{Full-system task completion experiments}
\label{tab:system}
\end{figure}

\subsubsection{Task 1} \label{sec:task1}

\begin{figure}[H]
    \centering
    \includegraphics[trim={2cm 0.5cm 2cm 1cm},clip,width=.49\linewidth]{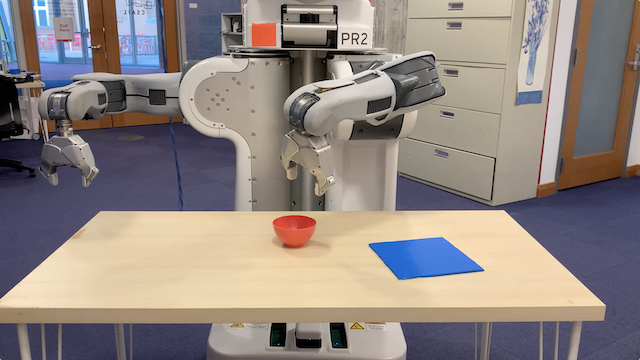}
    \includegraphics[trim={2cm 0.5cm 2cm 1cm},clip,width=.49\linewidth]{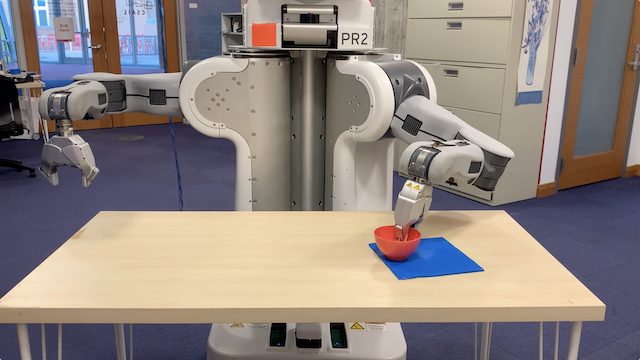}
    \caption{{\it Task 1}: the goal is for all objects to be on a blue target region. The robot picks up the bowl by grasping its interior and places it on the blue target region.
    }
    \label{fig:task1}
\end{figure}

This task evaluates our system's ability to grasp and stably place novel objects that are not well approximated by a simple box.
Almost all existing \tamp{} approaches assume that the manipulable objects can be faithfully modeled using a simple shape primitive for the purpose of manually specifying grasps.
The goal in this task is for all objects to be on a blue target region, which corresponds to the following logical formula: 
\begin{align*}
\forall {\it obj}.\; \exists {\it region}.\; \pred{On}({\it obj},{\it region}) \wedge \pred{Is}({\it region}, \val{blue}).
\end{align*}
In each trial, a single object is placed arbitrarily on the table. 
The five objects we used across the five trials were a bowl, a real power drill, a plastic banana, a cup, and a tennis ball.
Figure~\ref{fig:task1} demonstrates a successful trial where the object was a bowl.
A video of this trial is available at: \urlsmall{https://youtu.be/PREUU8nVetI}.

\subsubsection{Task 2} \label{sec:task2}

\begin{figure}[H]
    \centering
    \includegraphics[trim={2cm 0.5cm 2cm 1cm},clip,width=.49\linewidth]{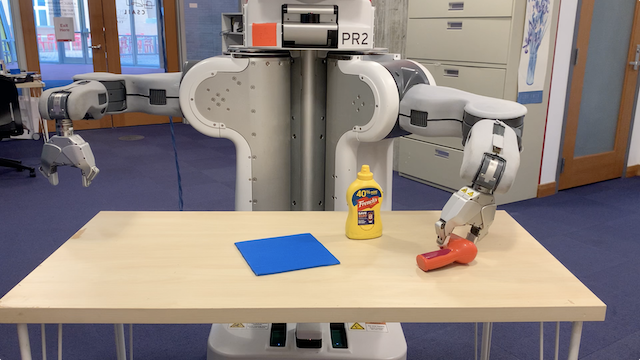}
    \includegraphics[trim={2cm 0.5cm 2cm 1cm},clip,width=.49\linewidth]{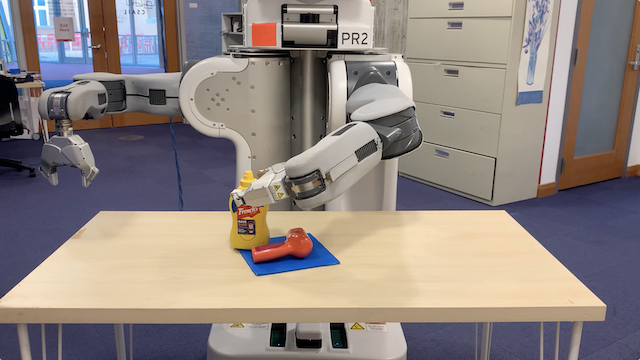}
    \caption{{\it Task 2}: the goal is for all objects to be on a blue target region.
    First, the robot first picks up the toy drill and places it near the bottom-right corner of the blue target region. Then, the robot picks up the mustard bottle and places it in the remaining collision-free area, which is near the top-left corner of the blue target region.}
    \label{fig:task2}
\end{figure}

This task evaluates our system's ability to safely place multiple objects in tight regions.
The goal in this task is also for all objects to be on a blue target region.
Two objects are initially present on the table, so the robot must plan a pair of placements and motions for the objects avoid collision. %
Figure~\ref{fig:task2} visualizes a successful trial involving a mustard bottle and a toy drill.
A video of this trial is available at: \urlsmall{https://youtu.be/BPa_Mpkf31M}.

\subsubsection{Task 3} \label{sec:task3}

This task evaluates our system's ability to react to unexpected observations.
The goal in this task is also for all objects to be on a blue target region. 
The task was presented in the introduction to the paper, see Figure~\ref{fig:task3_intro}. 

\begin{figure}[H]
    \centering
    \includegraphics[trim={1cm 0.5cm 3cm 1cm},clip,width=.49\linewidth]{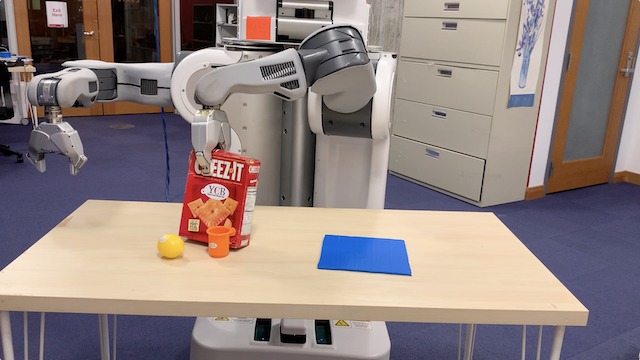}
    \includegraphics[trim={1cm 0.5cm 3cm 1cm},clip,width=.49\linewidth]{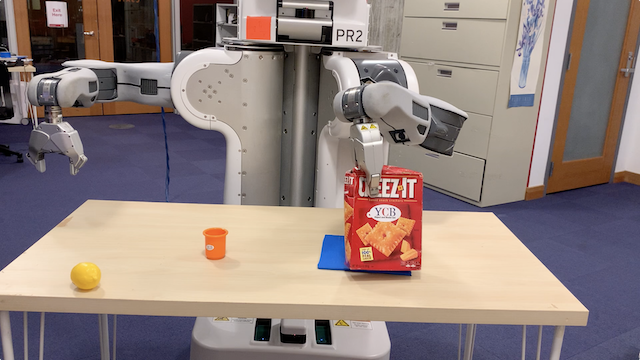} 
    \caption{{\it Task 3 failed trial}: the goal is for all objects to be on a blue target region. 
    The cracker box twists as the robot picks it up, which knocks the tennis ball outside the reachable workspace of the robot.
    }
    \label{fig:task3-failure}
\end{figure}

This task had the only failed trial among all five tasks.
Figure~\ref{fig:task3-failure} visualizes the failed trial.
The cracker box made contact with the occluded objects when lifted and knocked the tennis ball to the end of the table.
Upon re-observation, the robot identifies all three objects and deduces that the goal conditions are not satisfied. 
However, the robot fails to find a plan that achieves the goal within a generous timeout due to the fact that the tennis ball is now outside of the reachable workspace of the robot, causing the robot to fail to complete the task.
A video of this trial is available at: \urlsmall{https://youtu.be/NrTug_1EluI}.

\subsubsection{Task 4} \label{sec:task4}

\begin{figure}[H]
    \centering
    \includegraphics[trim={2cm 0.5cm 2cm 1cm},clip,width=.49\linewidth]{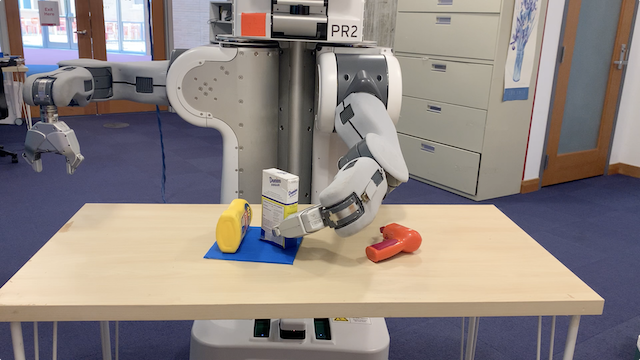}
    \includegraphics[trim={2cm 0.5cm 2cm 1cm},clip,width=.49\linewidth]{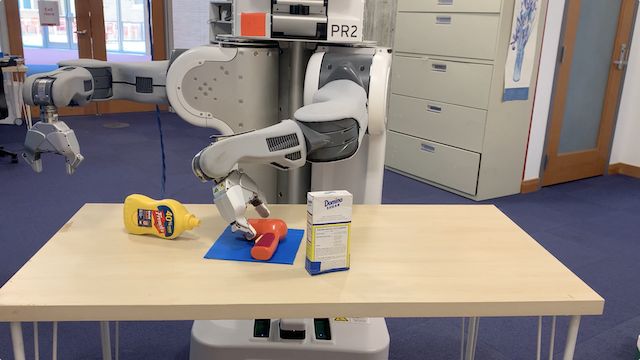}
    \caption{{\it Task-4}: the goal is for the object that is closest in color to red to be on a blue target region. 
    The robots picks and places both the sugar box and mustard box elsewhere in order to make room to place the toy drill, the most red object, on the goal target region.}
    \label{fig:task4}
\end{figure}

This task evaluates our system's ability to reason about collisions when placing a target object.
The goal in this task is for the object that is closest in hue to red to be on a blue target region, which corresponds to the following logical formula:
\begin{align*}
\exists {\it obj}.\; & \exists {\it region}.\; \forall {\it obj2} {\neq} {\it obj}.\;  \\ 
& \wedge \pred{On}({\it obj},{\it region}) \wedge \pred{Is}({\it region},\val{blue}) \\
& \wedge \pred{CloserInColor}({\it obj},{\it obj2},\val{red}).
\end{align*}
Initially, two objects that are far away in hue from red are located on the blue target region, occupying most of the region, and a third object that is close in hue to red is on the table. 
Figure~\ref{fig:task4} shows a successful trial. The robot picks and relocates a sugar bottle and mustard bottle that initially cover the blue target region in order to make room for the toy drill to be safely placed on the region.
A video of this trial is available at: \urlsmall{https://youtu.be/uqZT5gUBOo0}.

\subsubsection{Task 5} \label{sec:task5}

This task evaluates our system's ability to reason about collisions when attempting to pick objects.
The goal is for the object that is closest in color to yellow to be on a blue target region; the goal has a similar form to that in Task 4.
Initially, a mustard bottle is placed near the exterior of the table, surrounded by three potentially obstructing objects placed near the table's interior.

\begin{figure}[H]
    \centering
    \includegraphics[trim={2cm 0.5cm 2cm 1cm},clip,width=.49\linewidth]{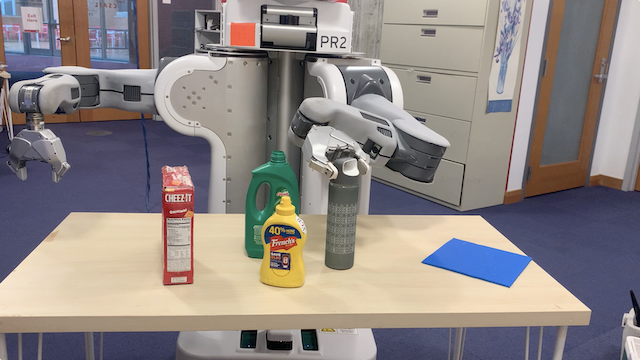}
    \includegraphics[trim={2cm 0.5cm 2cm 1cm},clip,width=.49\linewidth]{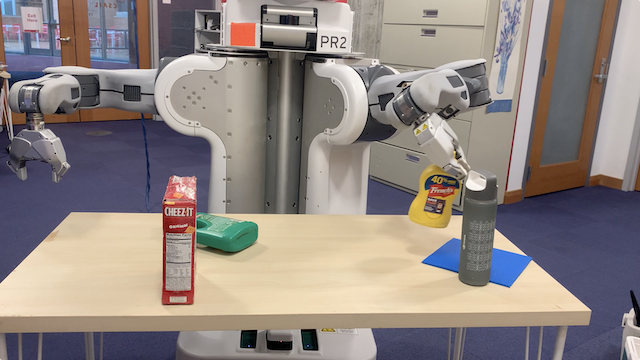}
    \caption{{\it Task 5}: the goal is for the object that is closest in color to yellow to be on a blue target region.
    The robot moves both the water bottle and detergent bottle out of the way in order to reach the mustard bottle and then place it on the blue target region.}
    \label{fig:task5}
\end{figure}

Figure~\ref{fig:task5} displays a successful trial.
First, the robot picks and relocates the obstructing water bottle. 
Second, the the robot picks and relocates the obstructing detergent bottle. 
It falls over during placement; however, the robot is able to infer this during its next observation and its estimates.
Finally, the robot picks the mustard bottle and places it on the blue target region.
A video of this trial is available at: \urlsmall{https://youtu.be/qBD2FyR2ktc}.

\section{Extensions to \name{}} %

\label{sec:extensions}
We have presented a simple %
instance of the \name{} framework, which is already quite capable, as illustrated by experimental results in Section~\ref{sec:system-experiments}.  It does have several assumptions and simplifications which can be removed, providing a path to even more general and capable systems.

\paragraph{Object features}

Object categories can play an important role in supporting the inference of latent object properties.  For example, recognizing that an object is likely to be an instance of the {\em coffee mug} category, based on its shape and appearance, might allow us to make additional inferences about its material, functional properties (can contain hot liquid), and parts (its opening and handle).
It is straightforward to augment the planning model %
so that if an object is perceived to have some property or class membership, then additional properties are inferred.  This capability enables examples, illustrated in Appendix~\ref{app:demos}, in which by recognizing an object as belong to the category {\em bowl}, we infer that objects can be dropped into it.

\paragraph{Additional manipulation operations} %
Extending the system to have additional manipulation operations is somewhat more complex, but the work is substantially amortized, again, over object arrangements, shapes, and goals. 
Added operations can be smoothly combined with the existing prehensile operation to generate a rich class of plans. 
For example, to add the ability to push an object, it would be necessary to add:  
\begin{itemize}
    \item a pushing controller to the robot description (although open-loop pushing could be accomplished with an existing position controller);
    \item a \pddlsmall{push} operator description that models the predicted change in object pose after a push;
    \item a \pddlsmall{plan-push} sampler, which can generate diverse choices of possible paths along which an object can be pushed, subject to some constraints which may include start and target poses.
\end{itemize}
Similar characterizations can be given for operations such as pouring and scooping~\cite{Wang2020LearningCM}, opening a child-proof bottle using impedance control~\cite{Holladay2021Planning},  moving kinematic objects such as drawers and doors~\cite{Garrett2019OnlineRI}, and many more.

\paragraph{State estimation}
In the basic system, there is no memory; actions are selected based only on the current view, which like many other vision-based manipulation approaches, is assumed to imperfect but sufficient for acting in the world.
For robustness, it is critical to integrate observations over time ({\it e.g.}, to remember objects that were once visible but are now not) and to integrate the predicted effects of actions ({\it e.g.}, to increase the belief that an object is located in a bowl after the robot drops it there, even if it cannot be observed inside the bowl from the current angle.
In addition, it can be beneficial to be able to fuse information from other sensory modalities, including tactile and auditory sensing.

\paragraph{Information gathering}
The current perception system generates a single hypothesis about the world state, which is used by the planner to select actions as if it were true.  %
Because it does not take into account the degree to which the robot is uncertain about the world state when it selects actions, it cannot decide that in some situations it would be better to do explicit information-gathering rather than pursue its goal more directly given a point estimate of the state.  
Previous work~\cite{Kaelbling2013IntegratedTA,HadfieldMenell2015ModularTA,Garrett2019OnlineRI} has provided methods for \tamp{} in belief space, but addressed uncertainty only in robot base and object instance pose, but not object shape or other properties.
Future work involves integrating these approaches with the proposed approach.

\section{Conclusion}

We have demonstrated an instance of a strategy for designing and building very general robot manipulation systems using a combination of analytical and empirical methods.  The system is a closed-loop policy that maps from images to position commands and generalizes over a broad class of objects, object arrangements, and goals.  It is able to solve a larger class of open-world sequential manipulation problems than methods that are either purely analytical (using classic hand-built algorithms for perception, planning, and control) or purely empirical (using modern methods for learning goal-conditioned policies).

\section*{Acknowledgements}

We gratefully acknowledge support from NSF grant 1723381; from AFOSR
grant FA9550-17-1-0165; from ONR grant N00014-18-1-2847; from the Honda Research
Institute; and from MIT-IBM Watson Lab. Caelan Garrett and Aidan Curtis are supported by NSF GRFP fellowships. 
Any opinions, findings,
and conclusions or recommendations expressed in this material are those of the authors and
do not necessarily reflect the views of our sponsors.

\clearpage

\bibliographystyle{IEEEtran}
\bibliography{references}

\clearpage

\input{appendix}

\end{document}

%% file: appendix.tex
\appendix

\subsection{PDDLStream} \label{app:pddlstream}

\begin{figure*}[h!]
\begin{footnotesize}
\begin{lstlisting}
(:action move-holding
 :parameters (?q1 ?t ?q2 ?oc ?g)
 :precondition (and (Motion ?q1 ?t ?q2) |\underline{(Grasp ?oc ?g)}| (AtGrasp ?oc ?g) (AtConf ?q1) 
                    (forall (?oc3 ?p3) (imply (AtPose ?oc3 ?p3) |\underline{(CFreeTrajGraspPose ?t ?oc ?g ?oc2 ?p2)}|)))
 :effect (and (AtConf ?q2) (not (AtConf ?q1)))
(:action pick
 :parameters (?q ?oc ?g ?p ?oc2 ?p2)
 :precondition (and |\underline{(Grasp ?oc ?g)}| (Kin ?q ?g ?p) |\underline{(Stable ?oc ?p ?oc2 ?p2)}|
                    (AtConf ?q) (HandEmpty) (AtPose ?oc ?p) (not (exists (?oc3) (On ?oc ?oc3))))
 :effect (and (AtGrasp ?oc ?g) (not (HandEmpty)) (not (AtPose ?oc ?p)) (not (On ?oc ?oc2))))
\end{lstlisting}
\end{footnotesize}
\caption{A \pddlstream{} description of \pddlsmall{move-holding} and \pddlsmall{place} actions. The underlined predicate is estimated by perceptual modules. %
} \label{fig:actions_all}
\end{figure*}

\begin{figure*}[h!]
\begin{footnotesize}
\begin{lstlisting}
(:stream predict-cfree-grasp
 :inputs (?t ?oc ?g ?oc2 ?p2)
 :domain (and (Traj ?t) (Grasp ?oc ?g) (Pose ?oc2 ?p2))
 :certified |\underline{(CFreeTrajPose ?t ?oc2 ?p2)}|)
\end{lstlisting}
\end{footnotesize}
\caption{A \pddlstream{} description of the \pddlsmall{predict-traj-grasp} stream, 
The underlined predicates denote properties estimated by perceptual modules.  
} \label{fig:streams_all}
\end{figure*}

This appendix provides the \pddlstream{} definitions for the remaining \name{} operators and streams that are not displayed in Figure~\ref{fig:actions} and Figure~\ref{fig:streams}.
Figure~\ref{fig:actions_all} gives the {\sc pddl} description of the \pddlsmall{move-holding} and \pddlsmall{pick} actions, which are similar to the \pddlsmall{move} and \pddlsmall{place} actions respectively.
Figure~\ref{fig:streams_all} gives the \pddlstream{} description of the \pddlsmall{predict-cfree-grasp} stream, which predicts whether object cloud \pddlsmall{?oc2} at pose \pddlsmall{?p2} does not collide with object cloud \pddlsmall{?oc} when held at grasp \pddlsmall{?g} while the robot follows trajectory \pddlsmall{?t}. A collision-free prediction certifies \pddlsmall{(CFreeTrajGraspPose ?t ?oc ?g ?oc2 ?p2)}.

\subsection{Perception module implementation} \label{app:perception}

This appendix provides details on the implementation and testing of the perception modules.

\subsection*{B.1 Segmentation}

Algorithm~\ref{alg:segment} displays the pseudocode for the combined \uois{} and DBSCAN segmentation algorithm.
It assumes a neural network for table segmentation $\id{NN}_t(\cdot)$ and a neural network for object segmentation $\id{NN}_s(\cdot)$, which can be implemented by \uois{}.
The algorithm's inputs are an RGB-D image $o$, the robot description ${\cal R}$ which contains the location of the RGB-D sensor, and the current robot configuration $q$.
It returns a set of segmented table point clouds $C_T$ and a set of segmented object point clouds $C_M$.

\begin{figure}[!h]
\centering \small
\begin{tabular}{lrrr} 
\toprule
Dataset & \dbscan & \uois & {\sc combined} \\
\midrule
{OCID Uncluttered} &
 \textbf{98.7} & 
 95.5 &
 97.2 \\
{OCID Cluttered} &
 68.7 &
 87.5 &
\textbf{89.6} \\
{GraspNet-1Billion} &
 69.1 &
 80.1 &
\textbf{82.6} \\
\bottomrule
\end{tabular}
\caption{Comparison of the segmentation approaches in terms of F-measure.}
\label{tab:seg}
\end{figure}

We compare all three segmentation methods on the ARID-20 subset of the {\it object clutter indoor dataset} (OCID)~\cite{Suchi2019easylabel} and {\it GraspNet-1Billion}~\cite{Fang2020graspnet} datasets, which contain several image distributions that vary object model complexity and the amount of clutter. The \textit{free} subset (390 images) of the ARID-20 dataset contains cases where objects are relatively well-separated from each other. The \textit{touching} and \textit{stacked} subsets (650 images in total) contain cluttered scenes that include objects with simple geometries.
The \textit{novel object} subset (7680 images) of the GraspNet-1Billion~\cite{Fang2020graspnet} dataset contains objects with geometries that are more complicated than simple boxes or cylinders. 960 images are sampled for testing purpose.
We used the default parameters for \uois{} a distance-from-neighbors parameter of $0.01$ meters for \dbscan{}.  The ground-truth non-object foreground mask was provided to \dbscan{} in order to segment out tables. For a fair comparison, we also provided the ground-truth foreground mask to \uois{} and the combined method as well.   The results are shown in Figure~\ref{tab:seg}. Results are reported using F-measure.

\begin{algorithm}[!h]
  \caption{The combined segmentation algorithm}
  \label{alg:segment}
  \begin{algorithmic}[1] %
    \begin{small}
    \Require Table segmentation network $\id{NN}_t(\cdot)$
    \Require Object segmentation network $\id{NN}_s(\cdot)$
    \Procedure{segment-cloud}{$o, {\cal R},  q$}
    \State $p_o \gets \Call{camera-pose}{{\cal R}, q}$
    \State $C_T \gets \{\Call{tform}{p_o, c_t} \mid c_t \in \id{NN}_t(o)\}$ \Comment{Table clouds}
    \State $c_T \gets \{v_t \mid c_t \in C_T, v_t \in c_t\}$ \Comment{Table points}
    \State $C_M \gets \emptyset$ \Comment{Object clouds}
    \For{$c_m \in \id{NN}_m(o)$}
        \State $c_m' \gets \Call{tform}{p_o, c_m} \setminus c_T$
        \State $C_M \gets C_M \cup \Call{dbscan}{c_m'}$
    \EndFor
    \State \Return $C_T, C_M$
    \EndProcedure
    \end{small}
\end{algorithmic}
\end{algorithm}

\subsection*{B.2 Surfaces}

We use the table segmentation predicted by \uois{} to identify the set of points to be considered in Random Sample Consensus (RANSAC)~\cite{fischler1981random} plane estimation.
Once the plane with the fewest outliers is found, we reclassify points in close proximity to the estimated plane as part of the table to account for segmentation errors.

Algorithm~\ref{alg:table} displays the pseudocode for the surfac3 estimation procedure.
It takes as input a segmented point cloud $c$, and outputs a surface $t = \langle p, h \rangle$ comprised by the pose of an $xy$-plane $p$ and a polygon $h$ within this plane.

\begin{algorithm}[!h]
  \caption{The surface estimation algorithm}
  \label{alg:table}
  \begin{algorithmic}[1] %
    \begin{small}
    \Require Orientation tolerance $\delta \approx 0$ in radians
    \Require Plane distance tolerance $\epsilon \approx 0$ in meters
    \Procedure{estimate-table}{$c$} \Comment{Point cloud $c$}
    \State $p \gets \Call{ransac}{c}$ \Comment{Plane pose $p$}
    \If{$(\Call{rot}{p} \cdot [0,0,1]) < \cos{\delta}$}
        \State \Return \kw{None}
    \EndIf
    \State $c_t \gets \Call{tform}{p^{-1}, c}$ \Comment{Local point cloud $c_t$}
    \State $b_t \gets \{[v_x, v_y] \mid v \in c_t, |v_z| \leq \epsilon\}$ %
    \State $h_t \gets \Call{convex-hull-2D}{b_t}$ \Comment{Polygon hull $h$}
    \State \Return $t = \langle p, h_t \rangle$
    \EndProcedure
    \end{small}
\end{algorithmic}
\end{algorithm}

\subsection*{B.3 Shape estimation}

Algorithm~\ref{alg:shape} displays the pseudocode for the version of the shape estimation algorithm that first combines the visible, MSN-predicted, and projected points, second filters visible points, and thirds returns a convex decomposition of the concave hull of the points.

\begin{figure}[!h]
\centering\footnotesize
\begin{tabular}{ llcccccc } %
\toprule
Clutter & Complex & %
\textit{V} & 
\textit{VP} & 
\textit{VPF} & 
\textit{VL} & 
\textit{VLF} & 
\textit{VLPF} \\ 
\midrule
No & No & 0.489 & 0.639 & 0.678 & 0.633 & 0.649 & \textbf{0.703}  \\
\hline
Yes & No & 0.417 & 0.556 & 0.589 & 0.565 & 0.573 & \textbf{0.612}  \\ 
\hline
No & Yes & 0.497 & 0.568 & 0.604 & 0.530 & 0.615 & \textbf{0.620}  \\
\hline
Yes & Yes & 0.398 & 0.464 & 0.483 & 0.446 & \textbf{0.503} & 0.497  \\ 
\bottomrule
\end{tabular}
\caption{Comparison of the shape-estimation strategies.}
\label{tab:sc}
\end{figure}

We experimented with the following six shape estimation strategies, all of which ultimately produce a mesh by taking the concave hull of one of following sets of points: 
a) the visible points only (\textit{V}), 
b) the union of the visible points and their vertical projection onto the table plane (\textit{VP}),
c) the points from \textit{VP} after filtering out points that are inconsistent with the observed depth image (\textit{VPF})
d) the union of the visible points and the learning based MSN shape-completed (\textit{L}) point cloud (\textit{VL}), 
e) the points from \textit{VL} after filtering (\textit{VLF}), and 
f) the filtered union of visible, projected, and MSN-predicted point clouds (\textit{VLPF}).

MSN was trained on a dataset comprised of 14 YCB object models rendered by PyBullet. %
Although we use YCB objects for training, we are aiming to learn a general object-shape prior rather than how to complete these specific objects. Testing is conducted on a much larger variety of objects.
Each object model was rendered individually from approximately 200 randomly-sampled view points. 
During training, the depth image was transformed to 3D space to produce the visible partial point cloud. 
The model was trained using the the Earth Mover's Distance (EMD) as the training loss function~\cite{liu2020morphing}.

We evaluated the performance of these methods in four different domains, each on 2000 images taken from a randomly-sampled camera pose. 
Each domain either contains imagines with a single object (\textit{No Clutter}) or with 10-14 objects (\textit{Clutter}).
The objects in each domain are either all primitive shapes (boxes, cylinders, and spheres) with randomized dimensions (\textit{Not Complex}) or a subset of 922 objects in 8 categories from the ShapeNet~\cite{shapenet2015} dataset, which include cameras, bowls, mugs, headphones,etc(\textit{Complex}).
To initialize the scene, objects are dropped sequentially on the table from a randomly-generated pose and simulated until arriving at a resting state. 
For scoring, the estimated meshes were converted to surface voxels and evaluated using boundary volumetric Intersection-over-Union (IoU). 
IoU scores close to one indicate that the estimated mesh covered most of the ground truth boundary voxels.
Boundaries were dilated to tolerate a small amount of misalignment that arises from the voxelization process.

Figure~\ref{tab:sc} shows the results of the experiment.
Using only the visible points (\textit{V}) performed the worst in each domain.
The fully combined method (\textit{VLPF}) in general performed the best across the domains, and filtering (\textit{F}) improved the accuracy of the predictions produced by both \textit{L} and \textit{P}.

Our experiments were conducted in scenes where objects are resting stably on a flat surface and the camera has an above viewing angle, which likely contributes to the projection approach (\textit{VP}) outperforming the learned approach (\textit{VL}) before filtering. The learned approach is a more general and flexible prior comparing to the projection method, which is a fixed heuristics. However the input(observed point cloud) will have small drift after going through the network, making the output inconsistent with the observed depth image. The filtering process can fix this problem so \textit{VLF} outperforms \textit{VPF} in complex domains.
In scenes where the top of an object could not be observed or the objects have complex geometries, we expect the learned approach to be more competitive.

\begin{algorithm}[!h]
  \caption{The combined shape estimation algorithm}
  \label{alg:shape}
  \begin{algorithmic}[1] %
    \begin{small}
    \Require Shape completion neural network ${\it NN_{sc}}(\cdot)$ %
    \Require Minimum number of points $n \geq 3$
    \Require Alpha shape parameter $\alpha$
    \Procedure{estimate-mesh}{$c, T, o, {\cal R}, q$} %
    \State $p_o \gets \Call{camera-pose}{{\cal R}, q}$
	\For{$\langle p_t, h_t \rangle = t \in T$}
        \State $c_t \gets \Call{tform}{p_t^{-1}, c}$
        \State $b_t \gets \{[v_x, v_y, 0] \mid v \in c_t, [v_x, v_y] \in h_t, v_z \geq 0\}$ %
        \State $c_t' \gets c_t \cup b_t \cup {\it NN_{sc}}(c_t)$ \Comment Completed point cloud $c_t'$
        \State $c_o'' \gets \{v \in \Call{tform}{p_o^{-1}p_t, c_t'} \mid v_z \geq o[v_r,v_c] \}$
        \If{$|c_o''| < n$}
            \State \kw{continue}
        \EndIf
        \State $m_o \gets \Call{concave-hull}{c_o'', \alpha}$ \Comment concave mesh $m_o$
        \State $M_o \gets \Call{convex-decomposition}{m_o}$
        \State $M \gets \Call{tform}{p_o, M_o}$ \Comment Convex meshes $M$
        \State \Return $M$
    \EndFor
    \State \Return \kw{None}
    \EndProcedure
    \end{small}
\end{algorithmic}
\end{algorithm}

\subsection*{B.4 Grasps}

\begin{algorithm}[!h]
  \caption{The EMA grasp generation algorithm}
  \label{alg:grasp}
  \begin{algorithmic}[1] %
    \begin{small}
    \Require Antipodal normal tolerance $\delta \approx 0$ in radians
    \Procedure{generate-grasps}{${\cal R}, m$}
    \State ${\cal E} \gets \proc{end-effector}({\cal R})$ \Comment{Extract the end effector}
    \While{\kw{True}}
        \State $v_1, \hat{n}_1 \gets \Call{sample-surface}{m}$ \Comment{Point $v_1$, normal $\hat{n}_1$}
        \State $v_2, \hat{n}_2 \gets \Call{sample-surface}{m}$
        \State $\hat{d}_{21} \gets {(v_2 - v_1)}/{|v_2 - v_1|}$
        \If{$-(\hat{n}_1 \cdot \hat{d}_{21}) < \cos{\delta}$ \kw{or} $(\hat{n}_2 \cdot \hat{d}_{21}) < \cos{\delta}$} %
            \State \kw{continue}
        \EndIf
        \State $\theta \gets \Call{sample-uniform}{0, 2\pi}$
        \State $R \gets \Call{AxisAngle}{\hat{d}_{21}, \theta}$
        \State $p \gets \Call{Pose}{R, (v_1 + v_2)/2}$ \Comment{Grasp pose $p$}
        \If{\kw{not} $\Call{collision}{{\cal E}, p, m, I_4}$}
            \State \kw{yield} $p$ \Comment{Generate the grasp}
        \EndIf
    \EndWhile
    \EndProcedure
    \end{small}
\end{algorithmic}
\end{algorithm}

Algorithm~\ref{alg:grasp} displays the pseudocode for the EMA grasp generator.
It takes as input the robot description ${\cal R}$ and an estimated mesh $m$.
It yields a possibly infinite sequence of grasp poses.
We performed a real-world experiment to compare the success rates of GPD, GraspNet, and EMA.
We used a simple definition of success where a grasp is successful if, after the robot performs a pick action with the grasp and then lifts its gripper, the robot is unable to fully close its fingers, which indicates that an object is in between the fingers.  Measuring the degree to which the object has changed pose relative to the robot would provide a more nuanced evaluation, but is beyond our current scope.  Each grasping trial was formulated as a call to the overall system with goal $\exists {\it obj}.\; \pred{Holding}(\val{left-hand},{\it obj})$.
In order to ensure that the planner does not always try to grasp the same object, we added a randomized cost to each individual pick action that biases the planner to select the object with lowest-cost feasible pick.
We performed two experiments, one in an {\em uncluttered} setting where objects are well-separated and one in a {\em cluttered} setting where objects are in contact with each other.
We performed 35 grasping trials in each setting. %

\begin{figure}[ht]
    \centering
    \includegraphics[trim={1cm 0.5cm 2.5cm 0.75cm},clip,width=.49\linewidth]{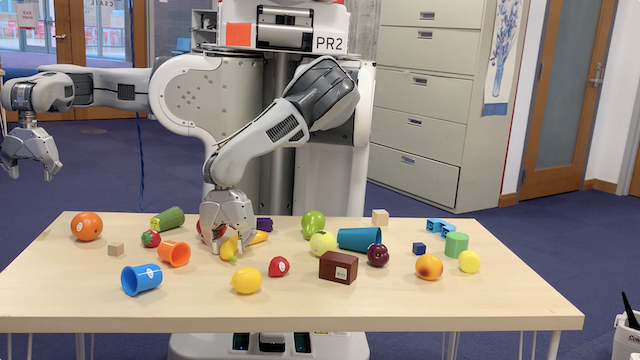}
    \includegraphics[trim={1cm 0.25cm 2.5cm 1cm},clip,width=.49\linewidth]{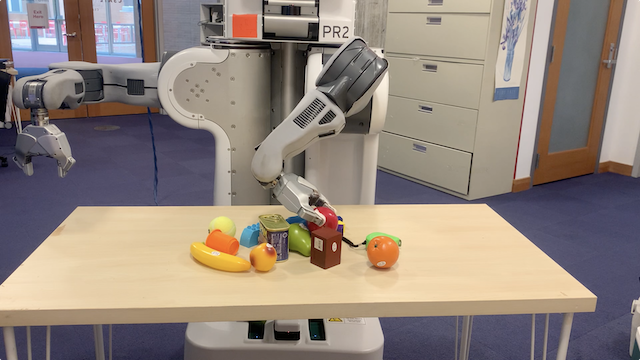}
    \caption{{\it Left}: a grasp generated by EMA in the {\it uncluttered} grasp experiment. {\it Right}: a grasp generated by EMA in the {\it cluttered} grasp experiment.}
    \label{fig:grasp_experiment}
\end{figure}

Figure~\ref{tab:uncluttered} shows the results of the grasp experiments, which are illustrated in Figure~\ref{fig:grasp_experiment}.
In the uncluttered case, the GPD and EMA methods both outperformed GraspNet and were able to produce a successful grasp on 27 out of 35 attempts.
A commonality between GPD and EMA is that they both identify candidate grasp transforms using geometric procedures, which operate either on a point cloud (GPD) or on a mesh (EMA), whereas GraspNet samples candidates directly from a neural network, which might not retain geometric invariants. %
GraspNet was also more computationally expensive than GPD and EMA, likely due to the inference overhead when using PointNet~\cite{qi2017pointnet}.

All three methods performed worse in the cluttered setting.  It makes segmentation more difficult, and as a result, both DBSCAN and \uois{} undersegment the pile, which causes multiple objects to frequently be included in a single instance point cloud.
Additionally, because objects are in close proximity to or even rest on top of others, some grasps that are viable for an object in isolation fail in clutter due to collisions with other objects and often transitively the rigid table, which alters the path of the robot's gripper.
Ultimately, EMA had a higher success rate than both GraspNet and GPD, possibly because it is able to consider object-gripper collisions during grasp generation since it operates on an estimated mesh, allowing it to prune grasps that push the target object during the grasping process.
Videos of the EMA experiments are available at: \urlsmall{https://youtu.be/VbKQmAvtRoY} and \urlsmall{https://youtu.be/mykqziKVrpI}.

\begin{figure}[ht]
\centering \footnotesize
\begin{tabular}{ lrrrr } \toprule
Method & \multicolumn{2}{c}{Uncluttered} & \multicolumn{2}{c}{Cluttered}\\
& {Successes} & {Time} & {Successes} & {Time} \\ 
\midrule
GPD & 27/35 & 8.7s & 23/35 & 6.8s \\ 
GraspNet & 21/35 & 48.3s & 16/35 & 21.7s \\ 
EMA ({\it ours}) & 27/35 & 10.3s & 27/35 & 7.3s \\ 
\bottomrule
\end{tabular}
\caption{
Grasp success rates in uncluttered and cluttered settings.
}
\label{tab:uncluttered}
\end{figure} %

\clearpage

\subsection{Demonstrations} \label{app:demos}

In addition to the multi-trial full-system experiments, we also tested our system with the following individual problems to qualitatively demonstrate the system's behavior; this appendix provides brief descriptions of these tasks.  See \urlsmall{https://tinyurl.com/open-world-tamp} for full videos of \name{} solving these problems.

\subsubsection{Packing three objects} %

\begin{figure}[H]
    \centering
    \includegraphics[trim={2cm 0.25cm 2cm 1.25cm},clip,width=.49\linewidth]{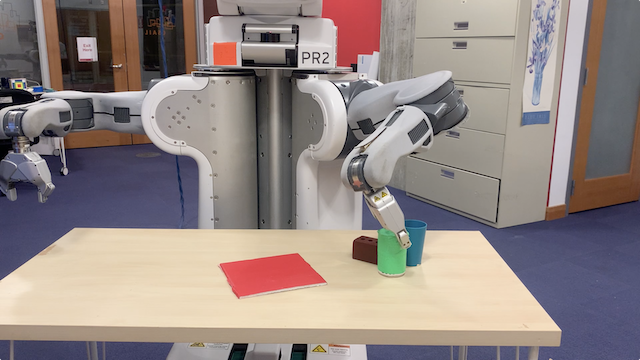}
    \includegraphics[trim={2cm 0.25cm 2cm 1.25cm},clip,width=.49\linewidth]{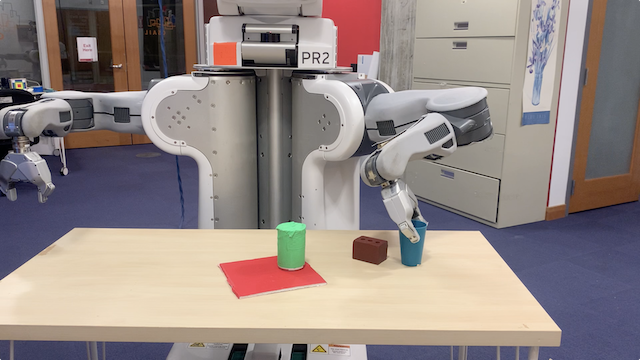} \\
    \vspace{4pt}
    \includegraphics[trim={2cm 0.25cm 2cm 1.25cm},clip,width=.49\linewidth]{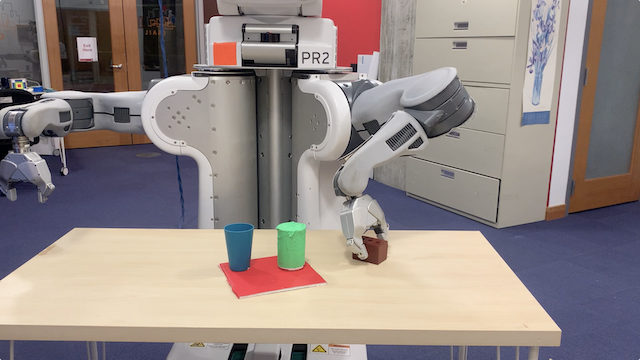}
    \includegraphics[trim={2cm 0.25cm 2cm 1.25cm},clip,width=.49\linewidth]{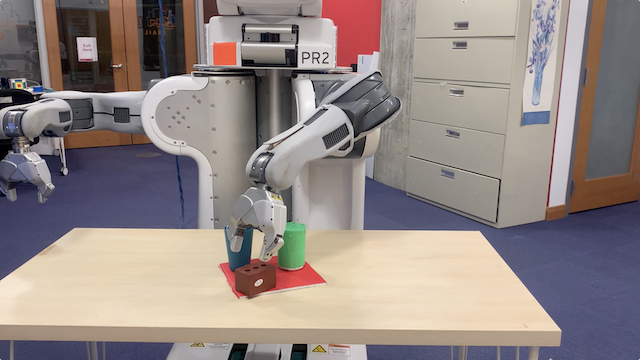}
    \caption{{\it Packing Three Objects:} the goal is for all objects to be on a red target region. 
    The robot finds placements for each of the three objects on the red target region that do not collide with each other.
    }
    \label{fig:brick}
\end{figure}

The goal in this task is for all objects to be on a red target region, which corresponds to the following logical formula:
\begin{align*}
\forall {\it obj}.\; \exists {\it region}.\; \pred{On}({\it obj},{\it region}) \wedge \pred{Is}({\it region}, \val{red}).
\end{align*}
Figure~\ref{fig:brick} displays still images from the video of our system solving this task, which is available at \urlsmall{https://youtu.be/qGbTOBECUPA}.

\subsubsection{Changing the resting face} %

\begin{figure}[H]
    \centering
    \includegraphics[trim={2cm 0.5cm 2cm 1cm},clip,width=.49\linewidth]{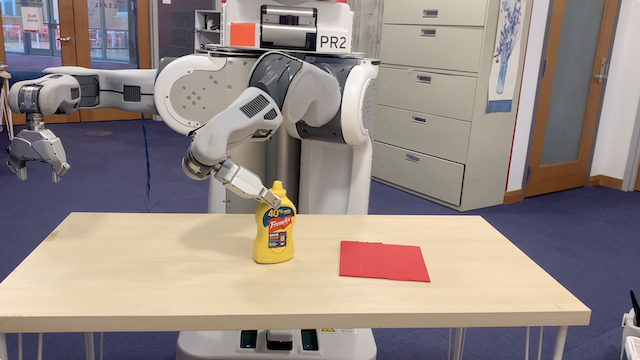}
    \includegraphics[trim={2cm 0.5cm 2cm 1cm},clip,width=.49\linewidth]{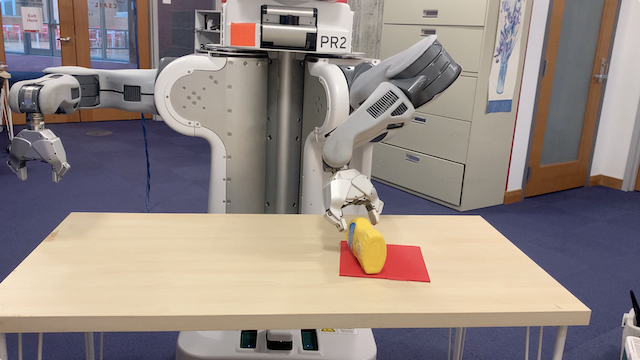}
    \caption{{\it Changing the Resting Face}: the goal is for all objects to be on a red target region.
    The robot identifies that the mustard bottle can be stably placed on its side and finds a plan with a placement that uses the mustard bottle's side as the resting face.
    }
    \label{fig:stable}
\end{figure}

The goal in this task is for all objects to be on a red target region, which corresponds to the following logical formula:
\begin{align*}
\forall {\it obj}.\; \exists {\it region}.\; \pred{On}({\it obj},{\it region}) \wedge \pred{Is}({\it region}, \val{red}).
\end{align*}
Figure~\ref{fig:stable} displays still images from the video of our system solving this task, which is available at \urlsmall{https://youtu.be/Er_m0CSsX7E}.

\subsubsection{Dispersing three objects} %

\begin{figure}[H]
    \centering
    \includegraphics[trim={2cm 0.5cm 2cm 1cm},clip,width=.49\linewidth]{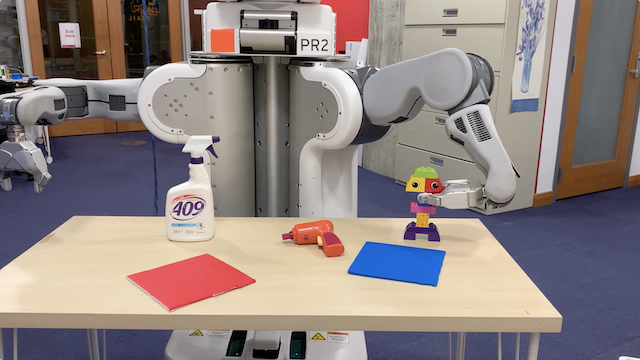}
    \includegraphics[trim={2cm 0.5cm 2cm 1cm},clip,width=.49\linewidth]{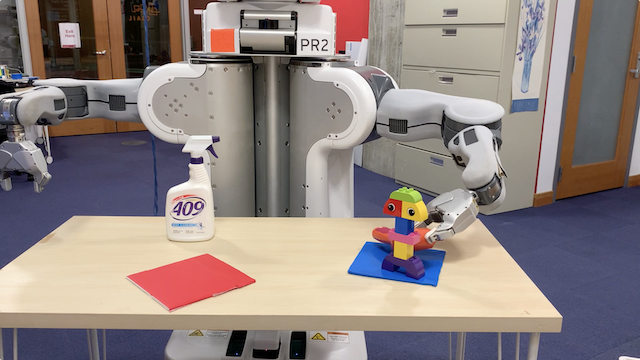} \\
    \vspace{4pt}
    \includegraphics[trim={2cm 0.5cm 2cm 1cm},clip,width=.49\linewidth]{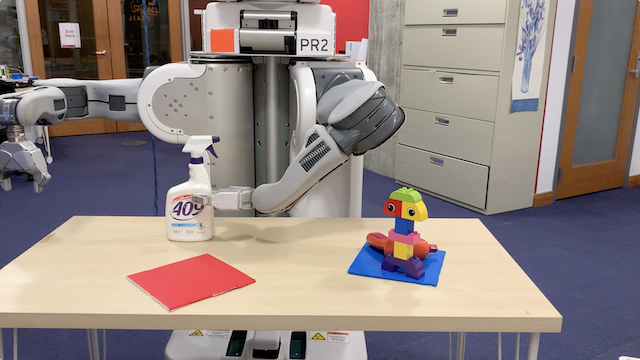}
    \includegraphics[trim={2cm 0.5cm 2cm 1cm},clip,width=.49\linewidth]{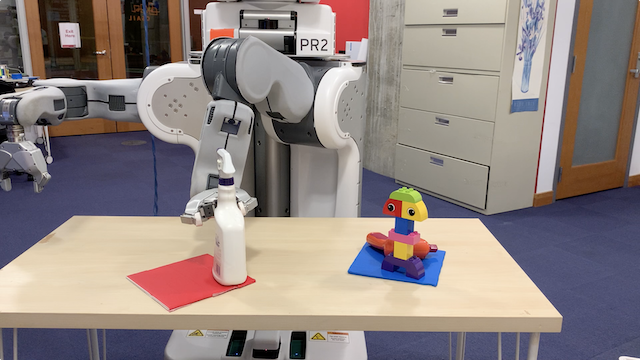}
    \caption{{\it Dispersing Three Objects:} the goal is for all objects to be on a target region. A red target region and blue target region are on the table.
    The robot plans to use both the red and blue target regions in order to have enough space to place all three objects.
    }
    \label{fig:lego}
\end{figure}

The goal in this task is for all objects to be on a target region, which corresponds to the following logical formula:
\begin{align*}
\forall {\it obj}.\; \exists {\it region}.\; \pred{On}({\it obj},{\it region}).
\end{align*}
In contrast to other tasks, both a red and a blue target region are on the table.
Figure~\ref{fig:lego} displays still images from the video of our system solving this task, which is available at \urlsmall{https://youtu.be/dWU8S8TuD3Y}.

\subsubsection{Manipulation in clutter} %

\begin{figure}[H]
    \centering
    \includegraphics[trim={2cm 0.25cm 2cm 1.25cm},clip,width=.49\linewidth]{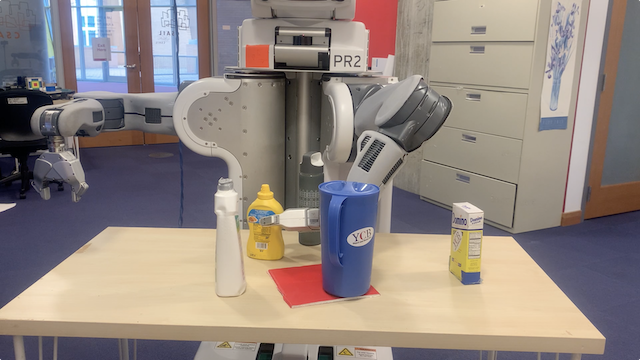}
    \includegraphics[trim={2cm 0.25cm 2cm 1.25cm},clip,width=.49\linewidth]{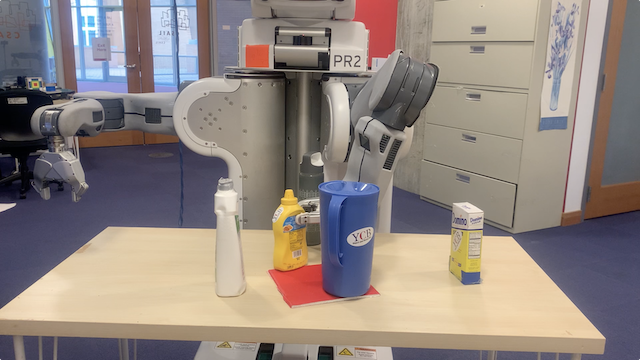}
    \caption{{\it Manipulation in Clutter}: the goal is for a mustard bottle to be on a red target region.
    The robot is able to carefully move between the four large objects to pick and place the mustard bottle while avoiding collisions.
    }
    \label{fig:traffic}
\end{figure}

The goal in this task is for a mustard object to be on a target region, which corresponds to the following logical formula:
\begin{align*}
\exists {\it obj}.\; & \exists {\it region}.\;  \pred{On}({\it obj},{\it region}) \wedge \pred{Is}({\it obj},\val{mustard})
\end{align*}
Figure~\ref{fig:traffic} displays still images from the video of our system solving this task, which is available at \urlsmall{https://youtu.be/UG8VOE1064A}.

\subsubsection{Occluded objects}

\begin{figure}[H]
    \centering
    \includegraphics[trim={2cm 0.5cm 2cm 1cm},clip,width=.49\linewidth]{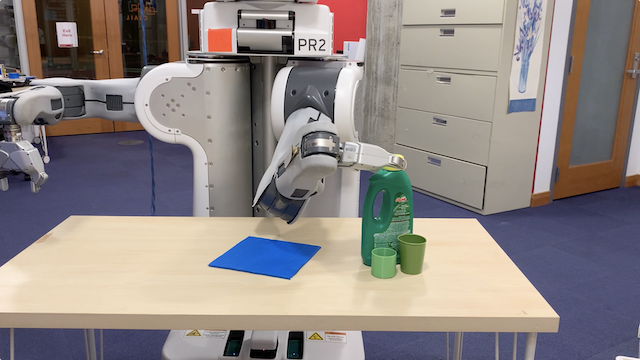}
    \includegraphics[trim={2cm 0.5cm 2cm 1cm},clip,width=.49\linewidth]{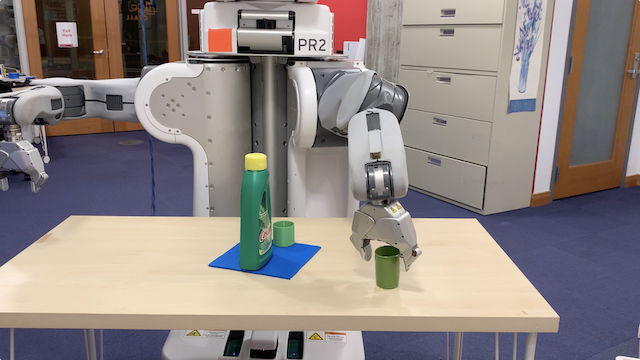} \\
    \vspace{4pt}
    \includegraphics[trim={2cm 0.5cm 2cm 1cm},clip,width=.49\linewidth]{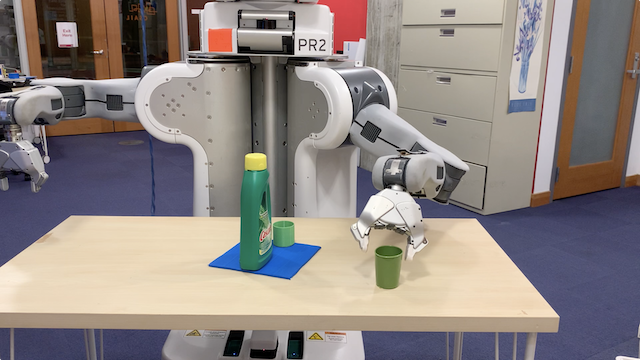}
    \includegraphics[trim={2cm 0.5cm 2cm 1cm},clip,width=.49\linewidth]{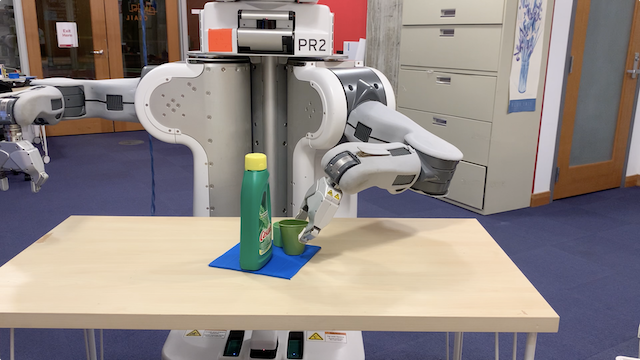}
    \caption{{\it Occluded Objects}: the goal is for all objects to be on a blue target region.
    }
    \label{fig:adian}
\end{figure}

The goal in this task is for all objects to be on a blue target region, which corresponds to the following logical formula:
\begin{align*}
\forall {\it obj}.\; \exists {\it region}.\; \pred{On}({\it obj},{\it region}) \wedge \pred{Is}({\it region}, \val{blue}).
\end{align*}
Figure~\ref{fig:adian} displays still images from the video of our system solving this task, which is available at \urlsmall{https://youtu.be/7x71aY2tc1M}.

\subsubsection{Obstructed place}

\begin{figure}[H]
    \centering
    \includegraphics[trim={2cm 0.25cm 2cm 1.25cm},clip,width=.49\linewidth]{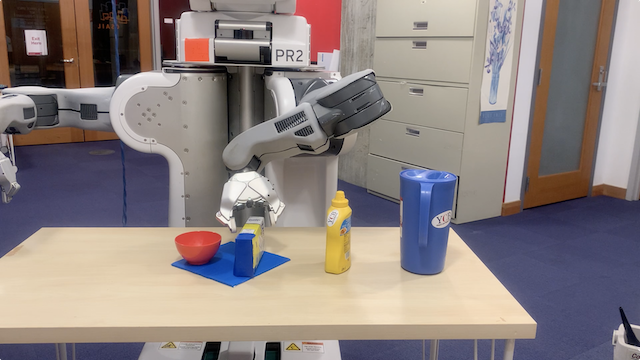}
    \includegraphics[trim={2cm 0.25cm 2cm 1.25cm},clip,width=.49\linewidth]{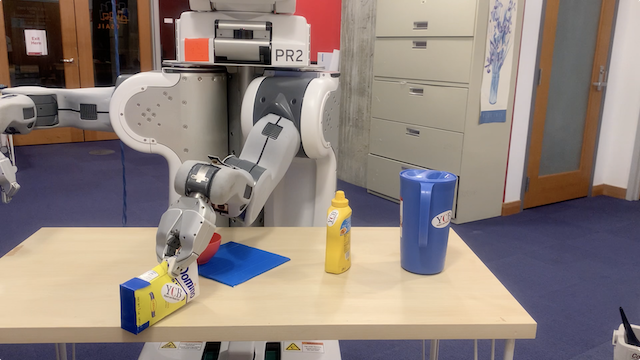} \\
    \vspace{4pt}
    \includegraphics[trim={2cm 0.25cm 2cm 1.25cm},clip,width=.49\linewidth]{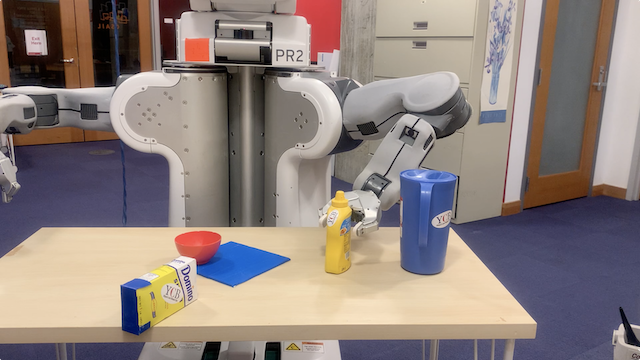}
    \includegraphics[trim={2cm 0.25cm 2cm 1.25cm},clip,width=.49\linewidth]{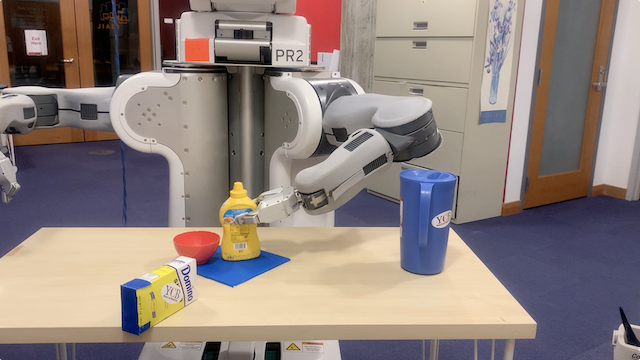}
    \caption{{\it Obstructed Place}: the goal is for a mustard bottle to be on a blue target region. The robot move the sugar box off of the blue target region to make room for the mustard bottle.
    }
    \label{fig:mustard}
\end{figure}

The goal in this task is for a mustard bottle to be on a blue target region, which corresponds to the following logical formula:
\begin{align*}
\exists {\it obj}.\; & \exists {\it region}.\;  \pred{On}({\it obj},{\it region}) \wedge \pred{Is}({\it obj},\val{mustard})
\end{align*}
Figure~\ref{fig:mustard} displays still images from the video of our system solving this task, which is available at \urlsmall{https://youtu.be/ipW0GAmM0yw}.

\subsubsection{Moving a destructible object} %

\begin{figure}[H]
    \centering
    \includegraphics[trim={2cm 0.5cm 2cm 1cm},clip,width=.49\linewidth]{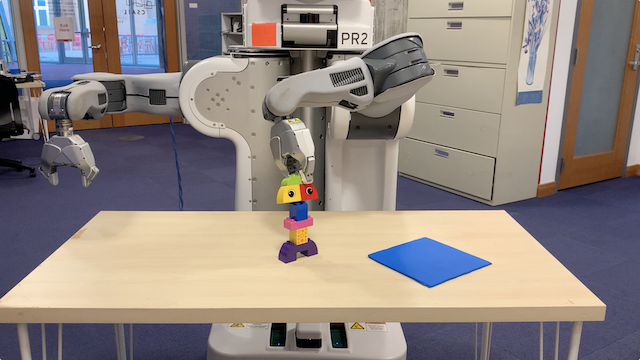}
    \includegraphics[trim={2cm 0.5cm 2cm 1cm},clip,width=.49\linewidth]{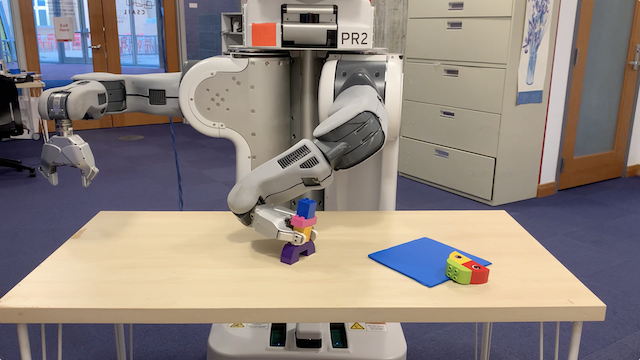} \\
    \vspace{4pt}
    \includegraphics[trim={2cm 0.5cm 2cm 1cm},clip,width=.49\linewidth]{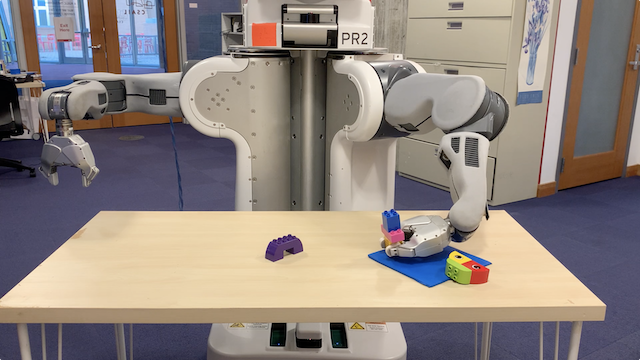}
    \includegraphics[trim={2cm 0.5cm 2cm 1cm},clip,width=.49\linewidth]{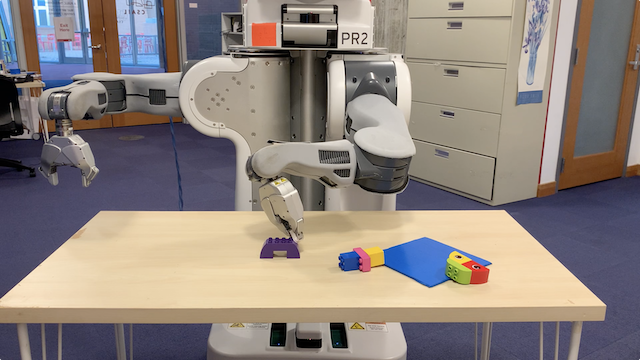} \\
    \vspace{4pt}
    \includegraphics[trim={2cm 0.5cm 2cm 1cm},clip,width=.49\linewidth]{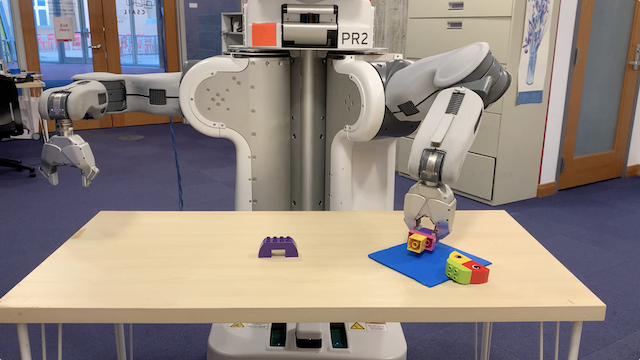}
    \includegraphics[trim={2cm 0.5cm 2cm 1cm},clip,width=.49\linewidth]{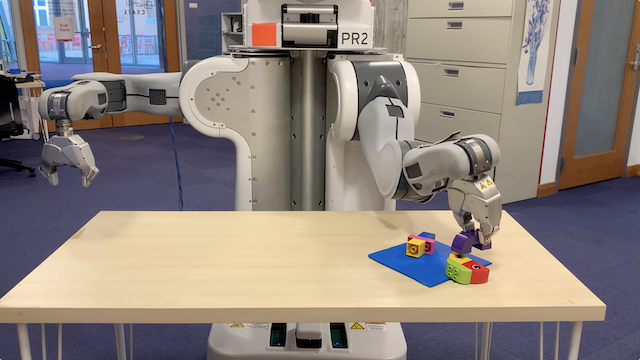}
    \caption{{\it Moving a Destructible Object}: the goal is for all objects to be on a blue target region.
    The Lego object is made up of a set of weakly-attached bricks and thus can break when manipulated.
    When the robot attempts to pick up the Lego, it breaks into three pieces, effectively creating two new objects.
    Still, the robot is able to perceive these new objects and re-plan to ensure that each one of them ends up on the blue target region.
    }
    \label{fig:destruction}
\end{figure}

The goal in this task is for all objects to be on a blue target region, which corresponds to the following logical formula:
\begin{align*}
\forall {\it obj}.\; \exists {\it region}.\; \pred{On}({\it obj},{\it region}) \wedge \pred{Is}({\it region}, \val{blue}).
\end{align*}
Figure~\ref{fig:destruction} displays still images from the video of our system solving this task, which is available at \urlsmall{https://youtu.be/ZpkzoQcxjW8}.

\subsubsection*{Re-grasp} %

\begin{figure}[H]
    \centering
    \includegraphics[trim={1cm 0.5cm 3cm 1cm},clip,width=.49\linewidth]{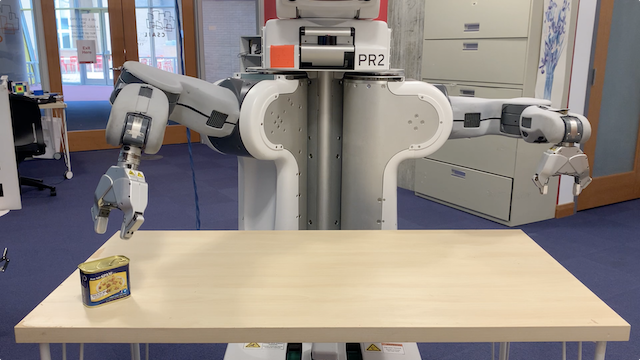}
    \includegraphics[trim={1cm 0.5cm 3cm 1cm},clip,width=.49\linewidth]{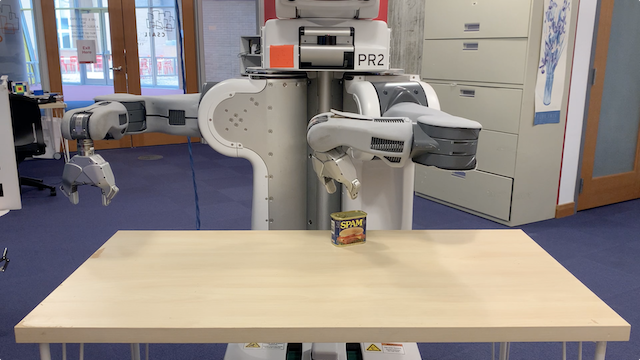}
    \caption{{\it Re-grasp}: the goal is for an object to be in the robot's left hand.
    The spam can is initially on the far right side of the table.
    Because its left arm is not able to initially reach it, the robot plans to pick up the spam with its right arm and place it in reach of the left arm.
    }
    \label{fig:regrasp}
\end{figure}

The goal in this task is for an object to be in the robot's left hand, which corresponds to the following logical formula:
\begin{align*}
\exists {\it obj}.\; \pred{Holding}(\val{left-hand},{\it obj}).
\end{align*}
Figure~\ref{fig:regrasp} displays still images from the video of our system solving this task, which is available at \urlsmall{https://youtu.be/J6VL7jbsonU}.

\subsubsection{Sorting} \label{sec:sort}

\begin{figure}[H]
    \centering
    \includegraphics[trim={2cm 0.5cm 2cm 1cm},clip,width=.49\linewidth]{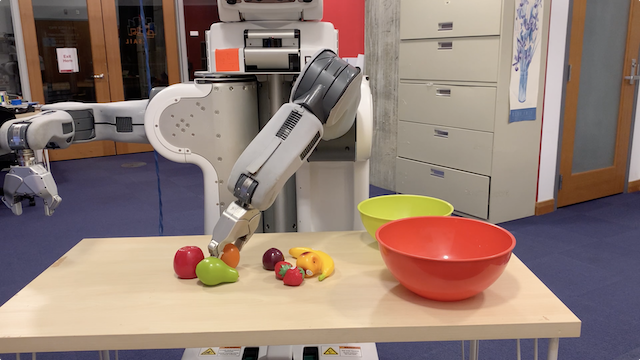}
    \includegraphics[trim={2cm 0.5cm 2cm 1cm},clip,width=.49\linewidth]{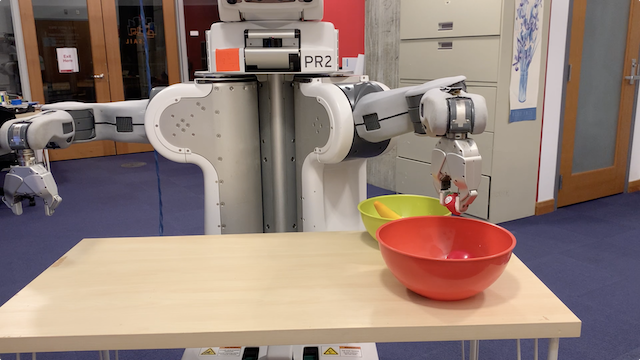}
    \caption{{\it Sorting}: the goal is for each object to be in the bowl that is closest in color to to them.
    The robot is able to successfully solve a long-horizon task that requires picking closely-placed objects.
    }
    \label{fig:sort}
\end{figure}

The goal in this task is for each object to be in the bowl that is closest in color to to them, which corresponds to the following logical formula:
\begin{align*}
\forall {\it obj}.\; & \exists {\it bowl}.\; \exists {\it color}.\; \forall {\it bowl2} {\neq} {\it bowl}.\; \\
& \pred{In}({\it obj},{\it bowl}) \wedge \pred{Is}({\it obj},{\it color})  \\
& \wedge \pred{CloserInColor}({\it bowl},{\it bowl2},{\it color})
\end{align*}
Figure~\ref{fig:sort} displays still images from the video of our system solving this task, which is available at \urlsmall{https://youtu.be/rjIvd9tRPpg}.

